\documentclass{pnastwo1}

\usepackage[utf8]{inputenc} % allow utf-8 input
\usepackage[T1]{fontenc}    % use 8-bit T1 fonts
\usepackage{microtype}      % microtypography

\usepackage{graphicx,amsfonts,amsmath,amssymb,color}
\usepackage{nima1}
\newcommand{\lt}[1]{\left\lVert#1 \right\rVert}

\begin{document}

%%%%%%%%%%%%%%%%%%%%%%%%%%%%%%

%% For titles, only capitalize the first letter
%% \title{Almost sharp fronts for the surface quasi-geostrophic equation}

% \title{Insert title here}

%% Enter authors via the \author command.
%% Use \affil to define affiliations.
%% (Leave no spaces between author name and \affil command)

%% Note that the \thanks{} command has been disabled in favor of
%% a generic, reserved space for PNAS publication footnotes.

%% \author{<author name>
%% \affil{<number>}{<Institution>}} One number for each institution.
%% The same number should be used for authors that
%% are affiliated with the same institution, after the first time
%% only the number is needed, ie, \affil{number}{text}, \affil{number}{}
%% Then, before last author ...
%% \and
%% \author{<author name>
%% \affil{<number>}{}}

%% For example, assuming Garcia and Sonnery are both affiliated with
%% Universidad de Murcia:
%% \author{Roberta Graff\affil{1}{University of Cambridge, Cambridge,
%% United Kingdom},
%% Javier de Ruiz Garcia\affil{2}{Universidad de Murcia, Bioquimica y Biologia
%% Molecular, Murcia, Spain}, \and Franklin Sonnery\affil{2}{}}

%\title{To train or not to train: deriving optimal weights in deep neural networks}
\title{Separation of time scales and direct computation of weights in deep neural networks}

\author{
Nima Dehmamy\affil{1}{
CCNR, Northeastern University, Boston 02115 MA, USA}
Neda Rohani\affil{2}{
IVPL, Northwestern University, Evanston, 60208 IL, USA}
\and
Aggelos Katsaggelos\affil{2}{}
}

\contributor{~%Submitted to Proceedings of the National Academy of Sciences of the United States of America
}

%% The \maketitle command is necessary to build the title page.
\maketitle

%%%%%%%%%%%%%%%%%%%%%%%%%%%%%%%%%%%%%%%%%%%%%%%%%%%%%%%%%%%%%%%%
\begin{article}

\begin{abstract}
%Starting from gradient descent equations, we show that in many deep neural networks, weights of low-lying layers may be estimated using principal component analysis.
Artificial intelligence is revolutionizing our lives at an ever increasing pace.
At the heart of this revolution is the recent advancements in deep neural networks (DNN), learning to perform sophisticated, high-level tasks.
However, training DNNs requires massive amounts of data and is very computationally intensive.
Gaining analytical understanding of the solutions found by DNNs can help us devise more efficient training algorithms, replacing the commonly used mthod of stochastic gradient descent (SGD).
We analyze the dynamics of SGD and show that, indeed, direct computation of the solutions is possible in many cases.
We show that a high performing setup used in DNNs introduces a separation of time-scales in the training dynamics, allowing SGD to train layers from the lowest (closest to input) to the highest.
We then show that for each layer, the distribution of solutions found by SGD can be estimated using a class-based principal component analysis (PCA) of the layer's input.
This finding allows us to forgo SGD entirely and directly derive the DNN parameters using this class-based PCA, which can be well estimated using significantly less data than SGD.
%We show here that this it may be possible to circumvent the expensive gradient descent procedure and derive the parameters of a neural network directly from properties of the training data.
%We show that, near convergence, the gradient descent equations for layers close{\color{red}r?} to the input can be linearized and become stochastic equations with noise related to the covariance of data for each class.
% We derive the distribution of solutions of these equations and discover that it is related to a ``supervised principal component analysis.''
We implement these results on image datasets MNIST, CIFAR10 and CIFAR100 and find that, in fact, layers derived using our class-based PCA perform comparable or superior to neural networks of the same size and architecture trained using SGD.
We also confirm that the class-based PCA often converges using a fraction of the data required for SGD.
Thus, using our method training time can be reduced both by requiring less training data than SGD, and by eliminating layers in the costly backpropagation step of the training.
%Additionally, these findings partially elucidate the inner workings of deep neural networks and allow us to mathematically calculate optimal solutions for some stages of  classification problems, thus significantly boosting our ability to solve such problems efficiently.
% In particular, we show that, with ReLU activation, the first layer is most likely to converge the earliest, and the weights are closely related to the principal components of input and their co-occurrence with different label classes.
% Our analysis can be used recursively to construct trained layers which can replace a few of the lowest layers in a deep neural network.
% We show the utility of this with one and two layers utilizing MNIST, CIFAR10 and CIFAR100 databases.
% When the labels are very restrictive, higher layers start diverging from simple PCA, which is label-agnostic, and weights become more fine-tuned to the labels.
% We show evidence for this when comparing the results of the networks trained on CIFAR10 and CIFAR100 datasets.
% This observation will allow for much faster training of a large class of deep learning problems.
% Our results also take a step in the direction of elucidating what the hidden layers are being trained to do.
\end{abstract}

%% When adding keywords, separate each term with a straight line: |

%% Optional for entering abbreviations, separate the abbreviation from
%% its definition with a comma, separate each pair with a semicolon:
%% for example:
%% \abbreviations{SAM, self-assembled monolayer; OTS,
%% octadecyltrichlorosilane}

% \abbreviations{}

%% The first letter of the article should be drop cap: \dropcap{}
%\dropcap{I}n this article we study the evolution of ''almost-sharp'' fronts

%% Enter the text of your article beginning here and ending before
%% \begin{acknowledgements}
%% Section head commands for your reference:
%% \section{}
%% \subsection{}
%% \subsubsection{}
 
%\section{Introduction}

\dropcap{A}rtificial Neural Networks\footnote{By neural network, we exclusively mean feed-forward perceptron models.} are an integral component of Artificial Intelligence and are recently experiencing a surge in popularity owing to their ability to perform more and more complex and abstract tasks.
While many factors, including cheaper computational power and big data, as well as better training algorithms \cite{goodfellow2016deep}, have contributed the revived interest in neural networks, one aspect has remained unchanged in the past fifty years: the use of gradient descent to train a neural network.
The major change in recent years has been the ability to train ``Deep Neural Networks'' (DNN), meaning a network with a large number of layers (tens and sometimes hundreds).
In DNN the number of layers is a major contributing factor to the computation time and the amount of data required for training.
While training within each layer can be parallelized using GPUs and other hardware, the ``backpropagation'' step\footnote{ ``Error Backpropagation'' is the term used for the usual chain rule used in computing the gradients of the cost function.} cannot, and needs to be performed sequentially.
Thus, the following question arises: Can the parameters of layers be derived analytically, without gradient descent?
Doing so efficiently would eliminate the need for backpropagation to that layer and would significantly reduce training time.

Given their complex structure and the large number of parameters, it is highly nontrivial how the training of a deep neural network can converge into ``good'' solutions.
Naively speaking, DNNs have a mathematical structure similar to a physical model called spin glass \cite{amit1985spin,barkai1990statistical,gardner1988space}, which is a system of binary nodes interacting over a random network.
Minimizing the cost function when training a DNN is equivalent to finding the ground state (minimum energy configuration) in a spin glass.
Most spin glasses, especially with random interactions, have a very rough energy landscape full of local minima, hence finding the low energy states becomes an NP hard problem.
Thus, the question is how do DNNs manage to converge to satisfactory solutions?

Deep networks are much more efficient than shallower networks \cite{telgarsky2016benefits} in certain classification tasks.
It is known that lower layers capture
general features in the data, and as we make the network deeper, higher layers become more abstract and specific to the label classes \cite{yosinski2014transferable}.
%Another aspect of DNNs that needs to be understood is the way they
%abstract the information contained in the data, i.e. break it down into the general and specific features mentioned above \cite{yosinski2014transferable}.
The best example is perhaps the activation pattern of filters in convolutional (ConvNet) layers \cite{lecun1995convolutional} trained for image classification.
As we consider deeper layers, the trained filters %when projected down to the input layer,
represent more and more complex and  high-level features in the data \cite{lecun2015deep} constructed by combining lower-level features.
%The filters in higher layers are combining filters of lower levels and that is why they represent more complicated objects.
But how a network, in which weights of all layers are being updated simultaneously, %for all layers
ends up choosing sensible abstract features at each level is also not fully understood.
Note that, while ``pre-training''  of layers \cite{bengio2009learning, schmidhuber1992learning} as auto-encoders contributed to the revived success of neural networks, many %successful DNNs currently used in image processing tasks
recent DNNs, such as AlexNet \cite{krizhevsky2012imagenet} or VGGNET \cite{simonyan2014very}, converge without a pre-training step, mainly due to availability of large amounts of training data.
%The large amount of training data available today seems to be sufficient for both convergence and building %of
%abstract hierarchies of features.

The goal of this paper is to gain analytical understanding of the distribution of solutions that SGD converges to in DNNs trained for classification tasks.
We start from the SGD equations and show that with an unbounded activation function, such as the Rectified Linear Unit (ReLU), the widely used --and most successful-- Glorot initialization \cite{glorot2010understanding} leads to a hierarchy in the rate of dynamics of different layers, with lower layers evolving much faster than higher ones.
We show that this ``separation of time-scales'' %{\color {red} inja mi2nim be ye ja reference bedim vase concept e separation of time-scales?}
allows SGD to solve layers from the lowest to the highest.
Additionally, examining SGD for each layer near convergence reveals that the dynamics of different label classes approximately decouple. %: during dynamics of lower layer, higher layers are  effectively static.
SGD for each class then becomes a familiar ``Langevin equation'' \cite{coffey2004langevin,kadanoff2000statistical}.
We derive the distribution of solutions for each class, showing that the optimal solution can be found using PCA on the input for that class.
We test these findings by creating pre-trained convolutional layers that replace existing layers in networks.
On MNIST \cite{MNIST}, CIFAR10 \cite{CIFAR10} and CIFAR100 \cite{CIFAR10} we observe that our fixed, pre-trained layers perform on par or superior to ConvNet.
% We will focus on image classification in formulating the problem.
% In the experiments, we show that early convolutional layers can be replaced by our result which is derived without back-propagation.
% The subset is based on which principal components are most relevant to the output classes.
% Finally, we implement our results and introduce a pre-trained network architecture and show that for the first and second layers its performance is in par with, or better than, convolutional layers in MNIST and CIFAR100.
% In CIFAR10, we find that the naive version of our architecture still performs better than a convolutional layer in the first layer, but performs worse in the second layer, which, we argue, is expected of the naive version of the analysis.

\section{A word on stochasticity and the learning dynamics}
The problem that we will focus on is classification: We wish to map a set of input vectors $X=(x_1,\dots,x_N)$ to output label vectors $Y=(y_1,\dots,y_N)$ (assumed to be ``one-hot'' vectors) using a nonlinear function $f(\theta; X) = Y$, where $\theta$ are the parameters (i.e. weights and biases).
The function $f(\theta;X)$ summarizes the action of a neural network with parameters $\theta$ on the data $X$.
We discuss the structure of the neural network below.
To assess the goodness of fit, we have a cost function $g(\theta,X,Y)$, which we write as $g[\theta]$.
The $\theta$ are assumed to be bounded.
The goal is to find $\theta$ such that the cost $g[\theta]$ is minimized.
$g[\theta]$ is also assumed to be smooth, except on a set of measure zero.
The standard method for training DNNs is %Stochastic Gradient Descent (
SGD.
%).
In SGD, data is processed gradually in mini-batches.
In early steps, %since only a sample of the data has been used to estimate $g[\theta] $,
the $\theta$ will have stochastic fluctuations of order $\sigma_\theta/\sqrt{N}$, with $N$ being the number of data inputs processed.
%At any step, the number of $N$ processed data points introduces a standard error {\color {red} on parameters of the model theta?} of $s/\sqrt{N}$, where $s$ is the sample standard deviation.
Given a cost function $g[\theta]$ with a set of parameters $\theta$ (weights $w$ and biases $b$), the standard error defines a \textit{resolution limit} for SGD at step $N$: local minima in the landscape of $g[\theta]$ whose widths are smaller than $\sigma_{\theta} /\sqrt{N}$, ($\sigma_{\theta}$ is the standard deviation of $\theta$) are invisible to SGD because stochastic fluctuations are large enough to make $\theta$ escape such local minima (SI Fig. \ref{fig:stderror-schema}).
An important consequence of this is that, much like temperature in simulated annealing \cite{kirkpatrick1987optimization}, standard error results in an effective ``smoothing'' of the energy landscape in SGD in the usual sense of ``Kernel Density Estimation'' \cite{epanechnikov1969non,wand1995kernel},  %(Fig.\ref{fig:stderror-schema}),
%At step $N$, only regions of the landscape of $g[w]$ whose width is larger than $2 \sigma_w /\sqrt{N}$ are resolvable by SGD.
%An important consequence of this is that like simulated annealing, SGD will escape narrow local minima with high $g[w]$ in early steps and
forcing it
to find a low energy (cost) local minimum in the \textit{smoothened landscape} of $g[\theta]$.
As SGD progresses, the standard error diminishes and narrow local minima become resolvable.
%This process does not make the optimization problem convex, but it can significantly reduce its complexity and eliminate high cost local minima.
\outNim{
\begin{figure}[h]
    \centering
    \includegraphics[width=.9\columnwidth]{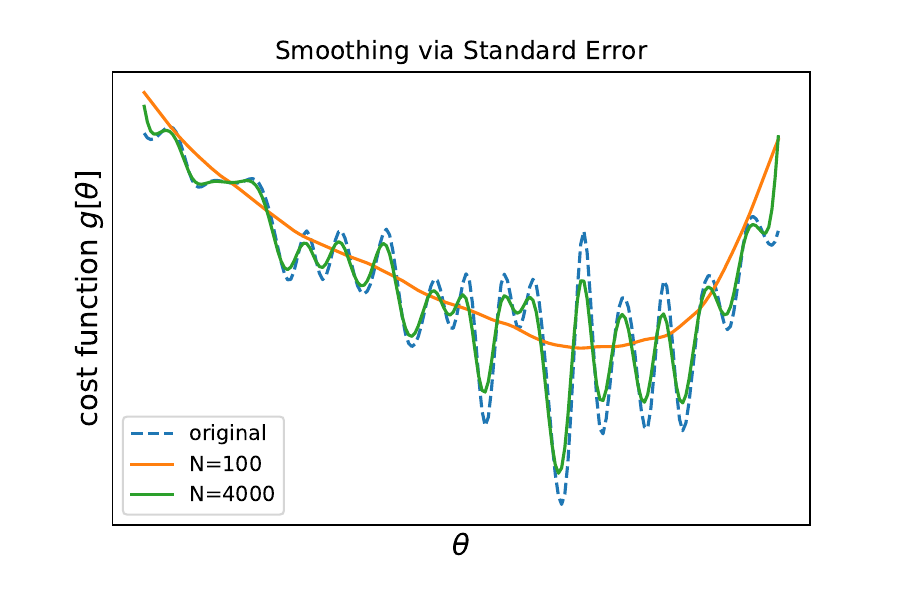}
    \caption{ {\bf Schematic of the smoothing effect of standard error on the landscape of the cost function $g[\theta]$. }
    At step $N$ the standard error in mean of parameters $\theta$ is $\ba{\sigma} = \sigma_\theta/\sqrt{N}$.
    Thus, stochastic fluctuations can move the $\theta$ by a normal distribution of width $\ba{\sigma}$.
    The effective energy landscape is then the convolution of the standard error distribution with the original $g[\theta]$ (dashed blue curve), resulting in a smoothing of the landscape at low $N$ (orange, $N=100$).
    At high $N$, the standard error is negligible and the original $g[\theta]$ is recovered (green $N=4000$).
    %{\color{red} what is g? the formulation also is W one dimension? range of w and g[w] on plot???}
    }
    \label{fig:stderror-schema}
\end{figure}
}%%%%%
Near the bottom of the smoothened minimum stochastic fluctuations dominate and the gradient becomes  negligible.
Thus, SGD will have two phases: 1) A ``fast drift phase'' driven by strong gradients, with a consistent direction; 2) A ``relaxation phase'' near the bottom of a local minimum\footnote{Note that these phases may occur multiple times as different layers may be entering this phase at different times and the landscape becomes less smooth during SGD.}.
These two phases were observed and utilized by \cite{shwartz2017opening} where the fast drift phase is described as a ``representation compression'' phase.
Fig. \ref{fig:sgd-phases} shows an actual SGD for a convex $g[\theta]$
to illustrate the two phases.  %{\color{red} again here we can mention what is the formulation of g}
Note that these two behaviors are a quite general feature of any stochastic process, such as diffusion, happening on an energy landscape with local minima.
\begin{figure}[h]
    \centering
    \includegraphics[width=\columnwidth, trim = 1cm 0 0cm 0,clip]{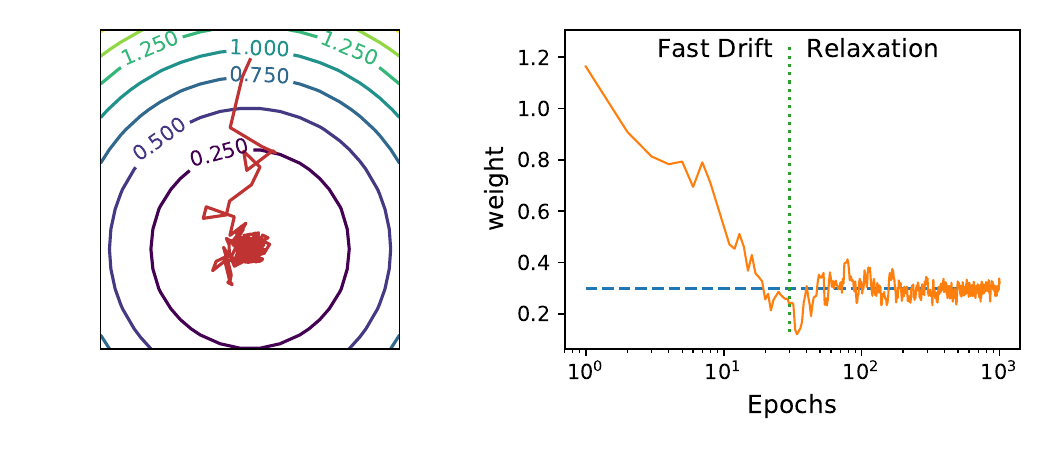}
    \caption{{\bf The two phase of Stochastic Gradient Descent (SGD):}
    The dynamics consists of two distinct phases: a fast drift phase, when the gradient of the cost function is large; a relaxation phase, where the gradient is negligible and stochastic fluctuations dominate the dynamics.
    In SGD, the spread of the fluctuations is due to statistical fluctuations in the input data used to estimate the cost function.
    As the number $N$ of training data increases, the spread of the fluctuations decreases like $\sigma_\theta /\sqrt{N}$ and the weight $\theta$ settles at the local minimum.
    %When Stochastic Gradient Descent (SGD) enters the vicinity of a local minimum, the dynamics will resemble diffusion in a potential well, except that the diffusion constant falls with increasing time steps.
    }
    \label{fig:sgd-phases}
\end{figure}
We show below that
the fast drift phase %almost certainly
can lead to a ``separation of time scales'': Layers closer to data have much faster dynamics than layers above them.
The condition for this to happen is having an activation function which is unbounded from above and a suitable initialization, e.g. Rectified Linear Units (ReLU) with Glorot initialization
\cite{glorot2010understanding,he2015delving}.
\outNim{
Most recent DNN architectures favor ReLU activation, defined simply as $f(x) = \max\{0,x\}$, over sigmoid or tanh for a number of reasons including lower computational complexity, constant gradient, faster learning,
and a reduced likelihood of vanishing gradient problem \cite{bengio1994learning, maas2013rectifier}.
It has also been shown that initial values of the weights affect the accuracy of DNNs significantly \cite{he2015delving}.
For ReLU, initialization methods such as variants of Xavier initialization, which limit the variance of weights to keep singular values of weights below two, work best \cite{glorot2010understanding,he2015delving}.
Our results suggest that one of  the reasons why ReLU with Xavier initialization works well is that the fast drift phase leads to weights acquiring singular values larger than one after a few steps, often even larger than two (Fig. \ref{fig:sv-growth}).
We show below that this leads to a separation of time scales in the dynamics of layers, with lower layers (close to input) changing significantly faster than layers above them.
This %separation of time scales
allows SGD to train the network effectively layer by layer.
}
When the system enters the relaxation phase, this separation of time scales still exists and we can utilize it to considerably simplify the SGD equations.
The relaxation phase is a
period where the gradients mostly vanish and the dynamics is dominated by stochastic fluctuations.
The system will be fine-tuning to find the exact position of the local minimum, as the details of landscape of $g[\theta]$ become resolvable.
It is important to note that each layer may enter the relaxation phase at a different time step.
\outNim{
During this phase, the layer is confined almost exclusively to a \textit{single} local minimum, as stochastic fluctuations are no longer large enough to allow it to tunnel to different minima.
Therefore, SGD in this phase becomes a \textit{convex optimization problem} for that layer and we can do a Taylor expansion of SGD around the local minimum.
}
% The expansion reveals that, similar to linear regression, SGD in the relaxation phase
% relates changes in weights $w$ to changes in the covariance of the input data, as we show below.
We will exploit this to analytically derive the distribution of weights. % which turns out to be a modified principal component analysis (PCA).
We will now delve into the details of the setup of the problem and the SGD equations.

\section{Setup and notation}
We consider the classification problem of labeling $N$ input images $X= (x_1,...,x_N)$ with $N$ one-hot vectors $Y= ({y}_1,...,{y}_N)$ as output labels.
We want to achieve this task using an $n$ layer neural network defined below.
The dimension of the output, or number of output ``channels'', of layer $k$ is denoted by $d^{(k)}$.
The labels $y_i$ have dimension equal to the number of classes $ C$.
%$= d^{(n)} $, which is equal to the output dimensions of the last layer.
%and $d^{(n)}=C$ is also the dimensions of each label vector $y_i$.
% Each layer of the network performs
The output of layer $k$ of the network is $h^{(k)} = f\pa{\tilde{h}^{(k)}}$ where $f(\cdot)$ is the ``activation function'' and $\tilde{h}^{(k)} = {w^{(k)}}^T h^{(k-1)} + b^{(k)}$, %is the  input to the activation function,
which we call the ``raw output''.
$w^{(k)}$ and $b^{(k)}$ are called the ``weights'' and ``biases'' of layer $k$, respectively.
$h^{(k)}_{ia}$ refers to channel $a$ of the output of layer $k$, for the $i$th input.
Thus, $h^{(0)}_i=x_i$ is the input and $h^{(n)}_i$ is the corresponding output of the network.
%{\color {red} age ma tu paper hich vaght az channel a estefade nemikonim aslan nemikhad inja ham biarimesh}

\begin{table}[h]
    \centering
    \caption{Notations}%
    \begin{tabular}{ll}
        Symbol & Notation\\
        \hline
         $n$ & \# of layers\\
         $N$ & \# of training data \\
         $C$ & \# of label classes \\%dimensions of input \\
         $d^{(k)}$ & dimension of layer $k$ output  \\
         $x_i$ & the $i$th input image  \\
         $y_i$ & label for input $i$th  \\
         $w^{(k)}$ & weights of layer $k$\\
         $b^{(k)}$ & bias of layer $k$, channel $a$\\
         $h^{(k)}_{ia}$ & output of layer $k$, channel $a$, image $i$\\
         \hline
    \end{tabular}
  \label{tab:notation}
\end{table}

\subsection{Activation Functions}
While in the past nonlinear bounded functions such as tanh and sigmoid were commonly used as activation functions,
most recent DNN architectures favor ReLU activation, defined simply as $f(x) = \max\{0,x\}$.
%over sigmoid or tanh
The preference for ReLU is due to a number of reasons including lower computational complexity, constant gradient, faster learning,
and a reduced likelihood of vanishing gradient problem \cite{bengio1994learning, maas2013rectifier}.
We will consider a feed-forward network,
%The setup of our network will be very minimal,
with the first $n-1$ layers having ReLU activation function\footnote{Note that, although ReLU is not a smooth function, it won't cause any problems in SGD because the non-smoothness is on a set of measure zero and discrete methods such as SGD will never discover the non-smooth part of the domain.
}.
The last layer is the classification layer with number of hidden nodes equal to the number of classes (i.e. $d^{(n)}=C$) and a softmax activation function defined below.
Table \ref{tab:notation} summarizes our notation. %Fig. \ref{fig:xay} sketches the structure of the system.
%In term $h_{ia}^{(k)}$, $i$ should not be confused with the channel index $a$.}
In most of what follows, we will only show the layer index $k$ and sometimes the input index $i$.
Matrix multiplication is implied unless stated otherwise.
%channel $a$ and location $z$ will usually be implicit in the matrix multiplications.
%In the past, sigmoid functions were commonly used as activation functions.
%However,
% Most recent DNN architectures favor ReLU activation, defined simply as $f(x) = \max\{0,x\}$, over sigmoid or tanh for a number of reasons including lower computational complexity, constant gradient, faster learning,
% and a reduced likelihood of vanishing gradient problem \cite{maas2013rectifier,bengio1994learning}.
With ReLU, the output $h^{(k)}$ of layer $k<n$ can be written as
\outNim{
\begin{equation}
h^{(k)}%\left(\tilde{h}^{(k)}\right)
= \theta\pa{\tilde{h}^{(k)}} \circ \tilde{h}^{(k)}, \qquad \tilde{h}^{(k)} = %w^{(k)}\cdot h^{(k-1)}
{w^{(k)}}^T h^{(k-1)}
+ b^{(k)}
\label{eq:act}
\end{equation}
}
\begin{equation}
h^{(k)}%\left(\tilde{h}^{(k)}\right)
= \mathrm{diag} \pa{ \theta \pa{\tilde{h}^{(k)}}} \tilde{h}^{(k)}, \quad \tilde{h}^{(k)} = %w^{(k)}\cdot h^{(k-1)}
{w^{(k)}}^T h^{(k-1)}
+ b^{(k)}
\label{eq:act}
\end{equation}
where $\theta(t)$ is the Heaviside step function and $\mathrm{diag}(f)$ is a diagonal matrix with $f$ on the leading diagonal.
%Here $\circ$ denotes element-wise multiplication\footnote{This can also be written as a matrix product using a diagonal matrix.}.
%In our analysis, we point out that choosing an unbounded activation function such as ReLU has another important consequence on the rate of dynamics of layers in the fast drift phase.
For the classification layer,
%, denoted as layer $n$,
we choose a softmax activation function, defined as $h_{ia}^{(n)} = \sigma(\tilde{h}^{(n)}_{ia}) = Z^{-1}_i\exp[\tilde{h}^{(n)}_{ia}] = \sigma_{ia} $, with the normalization factor being the sum over classes $Z_i = \sum_{a = 1}^{C} \exp[\tilde{h}^{(n)}_{ia}] $.

\subsection{Cost Function}
% \section{Cross-entropy}
% Suppose we have a set of labels $y_i$ and a Softmax activation
% \[h_i = {e^{z_i}\over Z}, \quad Z = \sum_j^C e^{z_j}. \]
In classification problems,
categorical cross-entropy provides a natural cost function for softmax activation function, as it measures the Kullback-Leibler divergence between the distribution of outputs $\sigma_{ia}$ and $y_{ia}$ labels%\footnote{that $y_i$ are one-hot vectors and only one of the categories is activated each time, meaning $\sum_a^C {y}_{ia} = 1$ {\color {red} In footnote mishe hazf she vazehe tu deep learning, age ja kam dashtim}.}
%Categorical cross-entropy provides a natural cost function for softmax activation function, based on likelihood maximization. %(Appendix \ref{ap:cross}
% \section{Cross-entropy\label{ap:cross}}
% We want a likelihood function that is maximized when for categories where $y_i=1$ the activation is also close to 1, and when $y_i = 0$, those terms don't play a role in the probability function.
% Thus we only want to maximize
%For each input image $i$, we want to maximize the likelihood function
%$P_i = \prod_{a=1}^C\sigma_{ia}^{y_{ia}} $
%where $a$ runs over the classes.
%This is equivalent to minimizing the cost function as the negative log-likelihood
\begin{align}
    {g}\pa{h^{(n)},{y}} & %=-{1\over N}\sum_{i=1}^N \log P_i
    =  -{1\over N}\sum_{i,a} {y}_{ia} \log \sigma_{ia} %\cr & =  - {1\over N}\pa{\sum_{i,a} {y}_{ia} \tilde{h}^{(n)}_{ia} -\sum_i\log Z_i}
    \label{eq:gcross}
\end{align}
%Gradient of \eqref{eq:gcross} yields
\outNim{
The extrema of \eqref{eq:gcross} are found from the roots of the gradient of the cost function as
\begin{align}
    {\ro {g} \over \ro \tilde{h}^{(n)}_i} &= {1\over N} \pa{ \sigma_i -{y}_i}=  {1\over N}\pa{{\exp[\tilde{h}^{(n)}_i] \over Z_i}-{y}_i }=0 \cr
    & \Rightarrow\tilde{h}^{(n)}_i = \log \pa{{y}_i Z_i}.
    \label{eq:dgdhn0}
\end{align}
%{\color{red}
Since the output of the activation functions we consider (such as ReLU, tanh or sigmoid) satisfy  $\tilde{h}_i^{(n)} > -\infty $, we have to regularize $y_{ia}$ so that the 0 values are replaced by $\min y_{ia} = \eps_i \sim 1/ Z_i$. %{\color{red} here we replace y which is 0 by 1/Z? log(yZ) = 0}
%}

%With this, we can define the weight-dependent variable ${y}_{ia} \equiv \log \pa{y_{ia} Z_i}$, for brevity.
%One of the key components in what follows will be obtaining a good estimate of what $y_{ia}$ should be.
%We return to this below.
} %% out

\subsection{Gradient Descent with ReLU\label{sec:gd}}
% \section{Depth and Gradient Descent}
Gradient descent consists of changing the weights and biases opposite to the gradient of the cost function to find minima.
The number of processed data points $N$ defines the time step of SGD in neural networks\footnote{During training, data points are also reused.
After one ``epoch'', i.e. after processing all training data once, the mini-batches are randomly sampled from the data set again.
For our purpose, we may treat reused data after the first epoch as new data.
We will not make a distinction between the different epochs.
}.
When a new mini-batch of size $\delta N$ is processed, the weights and biases will be changed according to
\begin{align}
    {\delta b^{(k)}\over \delta N}& = -\eps {\delta g\over \delta b^{(k)}}, &
    {\delta w^{(k)}\over \delta N}& = -\eps {\delta g\over \delta {w^{(k)}}^T}\label{eq:SGD}
\end{align}
where $\eps$ is the learning rate, which can be dynamically adjusted, and $\delta w \equiv w(N+\delta N)- w(N)$.
The components of gradient of the cost function are
\outNim{
\begin{align}
\delta g %&=  {\ro g \over \ro h^{(n)}}\delta h^{(n)} = {\ro g \over \ro w^{(n)}}\delta w^{(n)}+ {\ro g \over \ro b^{(n)}}\delta b^{(n)}+  {\ro g \over \ro h^{(n-1)}}\delta h^{(n-1)} \cr
&= \sum_{k=1}^{n-1} \pa{{\ro g \over \ro w^{(k)}}\delta w^{(k)}+ {\ro g \over \ro b^{(k)}}\delta b^{(k)} }. \label{eq:varg}
\end{align}
%where $k$ runs over the layers.
% \section{Gradient Descent with ReLU}
The partial variations of the weight and bias parameters for each layer are
}
\begin{align}
    {\delta g\over \delta b^{(k)}} & %={\ro g\over \ro  h^{(n)}} {\ro h^{(n)}\over \ro b^{(k)}}
    %= \pa{{\ro g\over \ro  \tilde{h}^{(n)}} }^T \prod_{m=k}^{n-1}{\ro \tilde{h}^{(m+1)}\over \ro \tilde{h}^{(m)}} {\ro \tilde{h}^{(k)}\over \ro b^{(k)}}\cr &
    = %{1\over N}\pa{ h^{(n)} - y}^T
    \pa{{\ro g\over \ro  \tilde{h}^{(n)}} }^T
    {A^{(k+1)}}^T, %\circ \theta\pa{\tilde{h}^{(k)}}^T
    & %\prod_{m=k}^{n-1}\theta \pa{ h^{(m+1)}} w^{(m+1)} \cr
    {\delta g\over \delta {w^{(k)}}^T}
    %& = {\ro g\over \ro  \tilde{h}^{(n)}} \prod_{m=k}^{n-1}{\ro \tilde{h}^{(m+1)}\over \ro \tilde{h}^{(m)}} {\ro \tilde{h}^{(k)}\over \ro {w^{(k)}}^T}\cr
    & =  h^{(k-1)} {\delta g\over \delta b^{(k)}}
    \label{eq:dgw}
\end{align}
\outNim{
\begin{align}
    {\delta g\over \delta b^{(k)}} & %={\ro g\over \ro  h^{(n)}} {\ro h^{(n)}\over \ro b^{(k)}}
    = \pa{{\ro g\over \ro  \tilde{h}^{(n)}} }^T \prod_{m=k}^{n-1}{\ro \tilde{h}^{(m+1)}\over \ro \tilde{h}^{(m)}} {\ro \tilde{h}^{(k)}\over \ro b^{(k)}}\cr &
    = %{1\over N}\pa{ h^{(n)} - y}^T
    \pa{{\ro g\over \ro  \tilde{h}^{(n)}} }^T
    {A^{(k+1)}}^T %\circ \theta\pa{\tilde{h}^{(k)}}^T
    \label{eq:dgb}\\ %\prod_{m=k}^{n-1}\theta \pa{ h^{(m+1)}} w^{(m+1)} \cr
    {\delta g\over \delta {w^{(k)}}^T}
    %& = {\ro g\over \ro  \tilde{h}^{(n)}} \prod_{m=k}^{n-1}{\ro \tilde{h}^{(m+1)}\over \ro \tilde{h}^{(m)}} {\ro \tilde{h}^{(k)}\over \ro {w^{(k)}}^T}\cr
    & =  h^{(k-1)} {\delta g\over \delta b^{(k)}}
    \label{eq:dgw}
\end{align}
}
\outNim{
\begin{align}
    {\delta g\over \delta b^{(k)}} & %={\ro g\over \ro  h^{(n)}} {\ro h^{(n)}\over \ro b^{(k)}}
    = \pa{{\ro g\over \ro  \tilde{h}^{(n)}} }^T \prod_{m=k}^{n-1}{\ro \tilde{h}^{(m+1)}\over \ro \tilde{h}^{(m)}} {\ro \tilde{h}^{(k)}\over \ro b^{(k)}}\cr &
    = %{1\over N}\pa{ h^{(n)} - y}^T
    \pa{{\ro g\over \ro  \tilde{h}^{(n)}} }^T
    {A^{(k+1)}}^T \circ \theta\pa{\tilde{h}^{(k)}}^T\label{eq:dgb}\\ %\prod_{m=k}^{n-1}\theta \pa{ h^{(m+1)}} w^{(m+1)} \cr
    {\delta g\over \delta {w^{(k)}}}
    %& = {\ro g\over \ro  \tilde{h}^{(n)}} \prod_{m=k}^{n-1}{\ro \tilde{h}^{(m+1)}\over \ro \tilde{h}^{(m)}} {\ro \tilde{h}^{(k)}\over \ro {w^{(k)}}^T}\cr
    & =  h^{(k-1)} \pa{{\ro g\over \ro  \tilde{h}^{(n)}} }^T %\pa{ h^{(n)} - y}^T
    {A^{(k+1)}}^T \circ
    \theta\pa{\tilde{h}^{(k)}}^T
    \label{eq:dgw}
\end{align}
}
where we have used the fact that $h \ro_h \theta(h) = h\delta(h) = 0$ and defined
\begin{equation}
    A^{(k)} \equiv \prod_{m=k}^{n}\mathrm{diag} \pa{ \theta \pa{ \tilde{h}^{(m-1)}} } w^{(m)}.
    %A^{(k)} \equiv \prod_{m=k}^{n}\theta \pa{ \tilde{h}^{(m-1)}} \circ w^{(m)}
    \label{eq:A}
\end{equation}
%where $\circ$ multiplies into each row of $w^{(m)}$.
% change during SGD  the depth $n$ of the network affects the convergence of SGD and the solutions it finds.
%We want to derive the condition for gradient descent to stop evolving for a particular layer.
We wish to understand the dynamics of \eqref{eq:dgw}.
The factors that depend on $k$ in \eqref{eq:dgw} are the weights of higher layers through $A^{(k+1)}$ and the output of the previous layer $h^{(k-1)}$.
We will show below that with ReLU the Frobenius norm of $A^{(k+1)}$ will be larger for smaller $k$.
For $h^{(k-1)}$, we will show that it will contain competing terms and together with the nonlinearity from ReLU, $h^{(k-1)}$ will not have a clear factor of weights and biases like $A^{(k+1)}$.

\section{Fast drift phase and singular values of $w^{(k)}$}
% As discussed earlier, because of standard $1/\sqrt{N}$ error, SGD cannot resolve sharp minima in early stages and the dynamics effectively occurs on a smoothened landscape of $g[w,b]$ with local minima that are much wider and far less abundant than in $g[w,b]$ without this smoothing effect.
% ReLU and the the intialization of weights play a role in the evolution of weights.

\begin{figure}[ht]
    \centering
    \includegraphics[width=1\columnwidth]{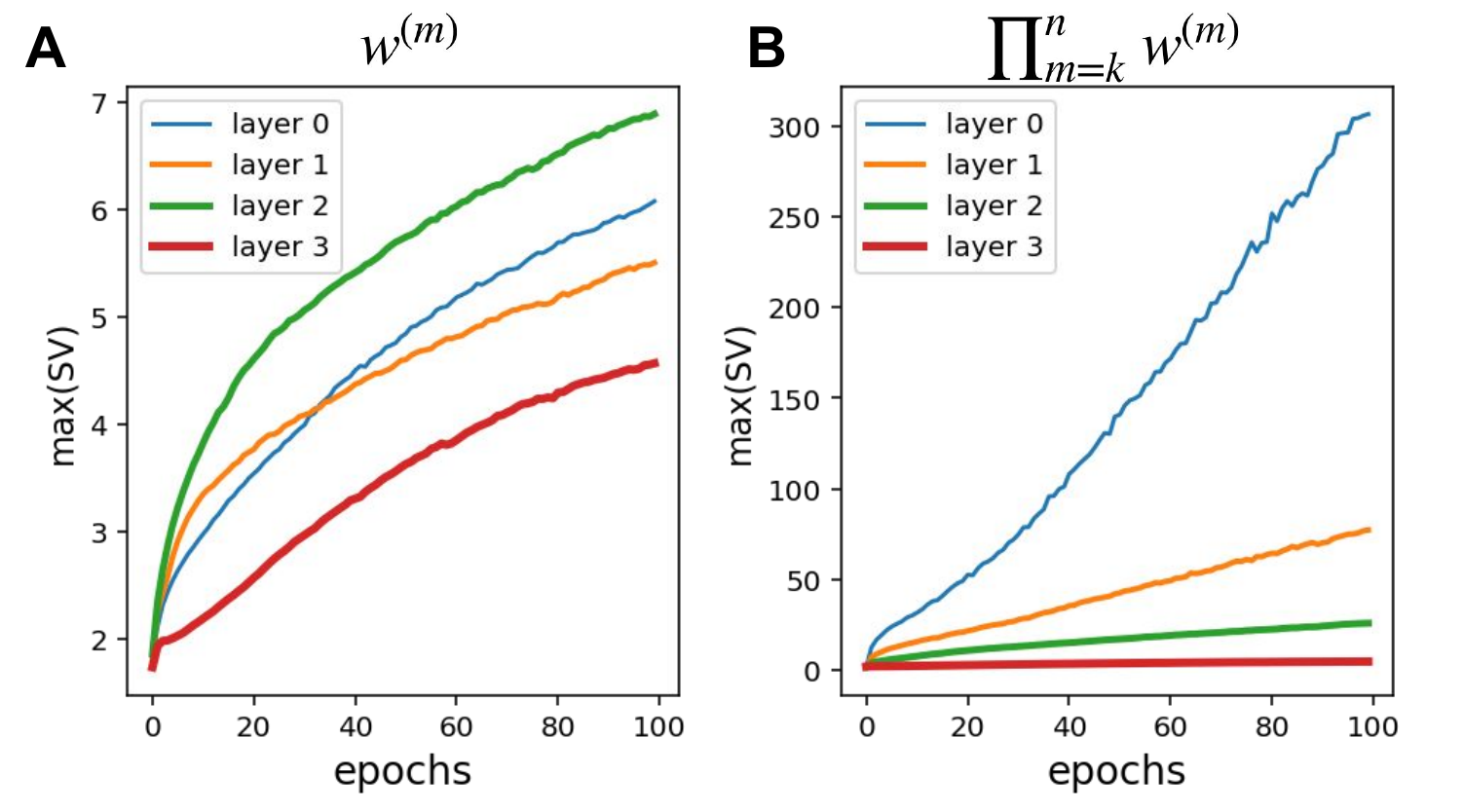}
    \caption{{\bf Singular Values and Growth of Gradients for Layers:} Using MNIST and a network with four fully connected layers (3 ReLU, one softmax classification, all using Glorot initialization).
    \textbf{A)} Although the initialization starts the networks with all singular values (SV) of all weights $w^{(m)}$ smaller than 2, the maximum SV quickly grows above 2.
    % {\color {red} inja title ro be jaye each layer bezarim $w^{m}$ vase weight haye har layer hast dge?}.
    \textbf{B)} The gradient for weights of layer $k$ is proportional to the product $\prod_{m=k}^n w^{(k)}$ of weights of all layers above it.
    As we predicted, the largest SV of the product of weights is greater for lower layers, thus pointing to faster dynamics of lower layers.
    }
    \label{fig:sv-growth}
\end{figure}

It is known that initial values of the weights affect the accuracy of DNNs significantly \cite{he2015delving}.
% For ReLU, initialization methods such as variants of Xavier initialization,
For ReLU, the best performing %\cite{he2015delving}
initialization scheme is found to be the Glorot method \cite{glorot2010understanding,he2015delving}
which limits the variance of weights so that the initial singular values of initial weights are equal to two. % \cite{glorot2010understanding,he2015delving}.
We will now show that Glorot initialization leads to a separation of time scales because it will result in $A^{(k)}$ acquiring singular values (SV) which are greater than $1$ and which are larger for smaller $k$.
Fig. \ref{fig:sv-growth} shows experimental evidence supporting this claim in a test network consisting of four dense layers\footnote{The exact architecture is: Maxpool (3,3), Dense(30), Dense(100), Dense(30), Classification(10).
The dense layers have about 3000 trainable parameters, while the classification layer has 310.} trained on the MNIST dataset.
It shows that $w^{(k)}$ of all layers quickly acquire SV $>1$ and that the product $\prod_{m=k}^n w^{(m)}$, as a proxy for $A^{(k)}$, has larger maximum SV for smaller $k$.
%Fig. \ref{fig:sv-growth} shows the growth of SVs, as well as the largest SV of the product of weight $\prod_{m=k}^n$
%It is in agreement with our argument that weights acquire SV larger than one (A) and that the largest SV of the product (B) is larger for lower layers.
In Glorot Initialization, all weights $w^{(k)}$ are initialized as random Gaussian (normal) distributions with variance ${\sigma}_{(k)}^2 =  2/d^{(k-1)}$.
This sets all SVs of $w^{(k)}$ to $\sqrt{2}$  (SI, eq. \eqref{eq:Mk}; see also \cite{glorot2010understanding}).

Now consider the combination $\tilde{w}^{(k)} \equiv \mathrm{diag} \pa{ \theta \pa{ \tilde{h}^{(k-1)}} } w^{(k)} $ which appears in $A^{(k)} = \prod_{m=k}^n\tilde{w}^{(m)}$.
Because the initialization is random Gaussian, the rows of the raw outputs $\tilde{h}^{(k)}$ are equally likely to be positive and negative, meaning half of the rows of $ \theta \pa{ \tilde{h}^{(k-1)}} $ are zero.
Therefore, the SVs of $\tilde{w}^{(k)}$ are initialized to $1$ and so all SVs of $A^{(k)}$ are also $1$ initially (SI eq.\eqref{eq:Mk}).
The SVs of $A^{(k)}$ are the square-root of eigenvalues of $B^{(k)}\equiv {A^{(k)}}^TA^{(k)}$.
In early steps, rows of $w^{(k)}$ are uncorrelated and so $B^{(k)}\approx 2^{k-n} \prod_{m=k}^n ||w^{(m)}||^2 I$ (SI, eq.\eqref{eq:Bk}).
\outNim{
To quantify the effect of $A^{(k+1)}$, we must examine its singular values and find out how they evolve with training steps, i.e. increasing $N$. % {\color{red} bebin tu har epoch hame point ha barresi mishan yani N sabete shayad step begi behtare. Chon jahaye ghabl ham az step estefade shode o ba epoch ye mani nemide}.
Define $\tilde{w}^{(k)} \equiv \mathrm{diag} \pa{ \theta \pa{ \tilde{h}^{(k-1)}} } w^{(k)} $.
At step zero, weights and biases of all layers are random and so we can assume that half of the entries of $\theta \pa{ \tilde{h}^{(k-1)}}$ will be zero, the other half one.
Thus, $\tilde{w}^{(k)}$ will have half of the rows of $w^{(k)}$ {\color{red}being nonzero we can remove this chon w ha ke sefr nistan wtilde sefr mishe, injuri bad khunde mishe}.
Consider the symmetric positive semi-definite $d^{(k)}\times d^{(k)}$ matrix $M^{(k)}\equiv \tilde{w}^{(k)T} \tilde{w}^{(k)}$.
The eigenvalues of $M^{(k)}$ are squares of singular values (SV) of $\tilde{w}^{(k)}$.
The $w^{(k)}$ are generally initialized randomly with zero mean.
As columns of $w^{(k)}$ are uncorrelated at the start, $M$ will be approximately diagonal and all diagonal entries will be similar as {\color{red} in tikke o Eq. ro tozih bede}
\begin{align}
    M_{ab}^{(k)} &= \sum_c \tilde{w}^{(k)}_{ca}\tilde{w}^{(k)}_{cb} \approx  {d^{(k-1)} \sigma^2_{(k)a}\over 2} \delta_{ab} \label{eq:Mk},%\cr
    % \sigma_{(k)i}^2 &= {1\over d^{(k-1)}} \sum_j \pa{\tilde{w}_{ji}^{(k)}}^2
\end{align}
where $\sigma_{(k)a}^2 = \sum_b \pa{w_{ba}^{(k)}}^2/ d^{(k-1)}$ is the variance of the weights over the input dimension and the $d^{(k-1)}/2$ is because of $\theta \pa{ \tilde{h}^{(k-1)}}$ eliminating half of the rows.
The variance of all rows is chosen to be the same $\sigma_{(k)a}^2=\sigma_{(k)}^2 $ initially.
Eq. \eqref{eq:Mk} implies that most eigenvalues of $M^{(k)}$ are initially close to $d^{(k-1)}\sigma_{(k)}^2/2$.
Now consider $B^{(k)} \equiv {A^{(k)}}^TA^{(k)}$ whose eigenvalues are squares of SV of $A^{(k)}$.
Using \eqref{eq:Mk} we have {\color{red} in tikke o Eq. ro tozih bede}
\begin{align}
    B^{(k)} & \approx {d^{(k-1)} \sigma^2_{(k)} \over 2}  {A^{(k+1)}}^TA^{(k+1)} \approx \prod_{m=k}^n {d^{(m-1)} \sigma^2_{(m)} \over 2} I
    \label{eq:Bk}
\end{align}
where
%$\ba{\sigma}^2_{(m)} = \sum_a \sigma^2_{(m)a}/d^{(k)}$ is the variance averaged over rows and
$I$ is the $C\times C$ identity matrix.
Experiments have shown \cite{he2015delving} that with ReLU, the Glorot initialization \cite{glorot2010understanding} (a variant of the Xavier method) which sets $\sigma_{(k)i}^2 = 2/ d^{(k-1)}$ yields better performance than other commonly used initialization methods.
Glorot initialization was designed specifically to set the maximum SV of all $w^{(k)}$ around one to avoid explosion of gradients, while choosing a smaller initialization makes the gradients too small and leads to worse results.
However, this initialization does not guarantee that the maximum SV will {\em remain} below one.
In fact, we argue that the reason this choice works better than a smaller initialization is precisely because the SV become larger than one early in SGD.
} %%%%%
As discussed earlier, because of standard error, which for $w^{(k)}$ at step $N$ becomes ${\sigma}_{(k)}/\sqrt{N}$, SGD cannot resolve sharp minima in early stages and the dynamics effectively occurs on a smoothened landscape of $g[w,b]$ with local minima that are wider and far less abundant than in $g[w,b]$ without this smoothing.
In early stages, $N \sim O(1)$ and, %the resolution of SGD for $w^{(k)}$ is $\ba{\sigma}_{(k)}$.
%The fast drift phase suggests that
therefore, the smoothened local minima %on the smoothened landscape of $g[w,b]$
have width $\Delta w^{(k)} \sim O\pa{{\sigma}_{(k)}}$.
Since in early stages, the gradient is not random and has non-zero magnitude, it maintains a consistent direction in the $w^{(k)}$ dimensions.
The gradient will, therefore, move the mean $\ba{w}^{(k)}_a = {1\over d^{(k-1)}} \sum_{b} w_{ba}^{(k)}$ of columns away from zero to $\ba{w}^{(k)}_a \sim O\pa{\sigma_{(k)}}$.
The largest SV of $\tilde{w}^{(k)}$ is the square-root of the largest eigenvalue of $\tilde{w}^{(k)T}\tilde{w}^{(k)} $ which in early stages with ${\sigma}_{(k)}^2 \approx  2/d^{(k-1)}$ yields %{\color {red} inja tozihat az ye jaie be ba'd ye kam gong mishe}
%$M^{(k)}$ then becomes
\begin{align}
    \mathrm{SV}_{\max{}} = \sqrt{{d^{(k-1)}\over 2}{\pa{ {\ba{w}^{(k)}_a}^2 +\sigma^2_{(k)}}} }%= {d^{(k-1)}\over 2}\pa{O(\sigma^2_{(k)}) +\sigma^2_{(k)}}
    > 1
\end{align}
%$\lambda_{\max{}} \approx d^{(k-1)}\pa{{1\over d^{(k)}} \sum_a{\ba{w}^{(k)}_a}^2 +\ba{\sigma}^2_{(k)}}/2$.
In other words, while the Glorot initialization is exactly tuned to set the {\em initial} SVs of $\tilde{w}^{(k)}$ equal to one, it also {\em guarantees} that the SVs become larger than one in early stages of SGD.
We will argue that this is essential in allowing SGD to solve layers one by one.
%Since Glorot initialization sets ${\sigma}_{(k)a} = 2/d^{(k-1)}$, SGD in the fast drift phase will certainly make $\lambda_{\max{}} > 1 $. %and closer to $\lambda_{\max{}} \approx 2$. {\color{red} ino daghigh tar tozih bede}
%Even if the variance diminishes significantly, this eigenvalue is very likely to become larger than $2$. {\color {red} in tikke be surate daghigh esbat nashode!}
%The conclusion is that in the fast drift phase with high likelihood, the weights of most layers will acquire one SV larger than two.
The immediate consequence of this  %and \eqref{eq:Bk} is that  of most %$w^{(k)}$ having an SV that is larger than one is that
is that $||A^{(k+1)}||$ will be larger for smaller $k$, as it contains $n-k$ factors of weights, each with some SVs larger than one. %close to $\sqrt{2}$.
Because of the multiplication of weights in $A^{(k)}$, the gradients of biases in \eqref{eq:dgw} are much larger for lower layers.
The weight gradients \eqref{eq:dgw} also contain another factor $h^{(k-1)}$, which we will discuss now.

\subsection{Magnitude of $h^{(k-1)}$}
Note that we cannot yet conclude that the norm of the weight gradients is larger for lower layers because of the $h^{(k-1)}$ factor in \eqref{eq:dgw}.
One may argue that $h^{(k)}$ contains the product of the weights of the first $k$ layers and therefore the gradients of $w^{(k)}$ for all $k$ are of the same magnitude.
But it is easy to see that $h^{(k)}$ cannot remain $ h^{(k)} \sim \prod_{m=1}^k w^{(m)} h^{(0)}$.
To see this, we examine the change in $\delta h^{(k)}$ after processing a $\delta N$ minibatch during SGD.
Using \eqref{eq:SGD}--\eqref{eq:dgw}, we have
\begin{align}
    % \delta h^{(k)} =& -\eps \pa{{\ro g \over \ro \tilde{h}^{(n)}} }^T
    % {A^{(k+1)}}^T \pa{1+ ||  h^{(k-1)}||^2} \cr
    {\delta \tilde{h}^{(k)} \over \delta N}
    %=& -\eps \left[ \pa{1+ |h^{(k-1)}|^2} A^{(k+1)} {\ro g \over \ro \tilde{h}^{(n)}}
    %+ {w^{(k)}}^T {\delta h^{(k-1)} \over \delta N} \right]\cr
    = & -\eps \sum_{m=1}^{k-1} \pa{1+ \lt{h^{(k-m)}}^2} W^{(m)T}W^{(m)}  A^{(k+1)} {\ro g \over \ro \tilde{h}^{(n)}}\cr W^{(m)}\equiv &\prod_{p=k-m+2}^k \mathrm{diag} \pa{ \theta \pa{ \tilde{h}^{(p-1)}} } w^{(p)}
    \label{eq:dhk}
\end{align}
Thus, SGD will change $\tilde{h}^{(k)}$  by terms proportional to $A^{(k+1)}$.
Therefore $h^{(k-1)}$ in  \eqref{eq:dgw} will not have a well-defined factor of weights and, in particular, $h^{(k)}$ cannot consistently remain close to $\prod_{m=1}^k w^{(m)} h^{(0)}$.

In conclusion, the only factor of weights that is guaranteed to be present in the evolution of weights (and biases) is the $A^{(k+1)}$ in \eqref{eq:dgw}.
As we showed above, $A^{(k)}$ has larger maximum SV for smaller $k$, resulting in lower layers having gradients which are larger than the layer above them by a factor larger than one. % close to $\sqrt{2}$ in early stages.
The significance of this is that it signals the existence of a ``separation of time scales'': the dynamics of lower layers is much faster than the layers above them.
This is a point worth deliberating because it means that the dynamics of layers approximately decouple and SGD in layer $k$ may effectively ignore dynamics of layer $k+1$, allowing SGD to %  the dynamics to efficiently
solve lower layers without much disruption from higher layers.
We will exploit this observation to solve the distribution of weights analytically.

In the end, note that both
ReLU and the initialization of weights played a role in this result.
Sigmoid or tanh activation %which were popular in the past but have fallen out of favor because of numerous issues including vanishing gradients and poor performance,
would not have resulted in growing SVs and the time scale of evolution of all layers would have been similar.
Additionally, the Glorot initialization seems to be the only initialization for ReLU which both makes sure the output of the layer is not exploding initially, while also being guaranteed to lead to SVs larger than one.

\section{Relaxation phase}
For each layer, when the parameters are in the vicinity of a local minimum, the fast drift phase ends and the layer enters the stochastic relaxation phase.
% The dynamics of the layer becomes a stochastic process at the bottom of a local minimum.
This phase sets in when weights are trained to a good degree and so the gradients $\delta g/\delta b^{(k)}$ and $\delta g/\delta w^{(k)}$ become very small.
From Eq. \eqref{eq:dgw}, this means that $\lt{\ro g/ \ro \tilde{h}^{(n)}}$ must be small.
% Since the weights and biases are close to convergence, $|h^{(k-1)}|$ becomes large enough in this phase, because the network is able to identify features of the data.
% Thus $\ro g/\ro \tilde{h}^{(n)}$ must become small
Using \eqref{eq:gcross}, for input $i$
\begin{align}
    \lt{{\ro g\over \ro \tilde{h}^{(n)}_i}}% & =  - {1\over N}\pa{\sum_{i,a} {y}_{ia} \tilde{h}^{(n)}_{ia} -\sum_i\log Z_i} \cr
    &= {1\over N} \lt{h^{(n)}_i-y_i} \ll 1
    \label{eq:dgdh-relax}
\end{align}
Since the gradient is very small, we can expand the exponential inside $h^{(n)}$ as a Taylor series\footnote{Note, $N$ can be large, but it's always finite.}.
First, we make the following $h^{(n)}$-dependent variable redefinition
\begin{equation}
    \tilde{y}_i \equiv \log \pa{y_i  Z_i}.
\end{equation}
where we replace the zeros in $y_i$ with a small positive $\eps \sim 1/Z_i$ to make $\tilde{y}_i$ well-defined.
%Note that this variable {\color {red} $\tilde{y}_i$?} redefinition is $h^{(k)}$-dependent. %, but as we are in the relaxation phase and changes in weights and biases are small, $Z_i$ is approximately the same function of input $h^{(0)}_i$
Define the projection matrix onto class of $y_i$ as $P_i \equiv \mathrm{diag}\pa{y_i}$.
\outNim{
\begin{equation}
    P_i \equiv \mathrm{diag}\pa{y_i}
    \label{eq:proj}.
\end{equation}
}%%%
Expanding \eqref{eq:dgdh-relax}, we have
\begin{align}
    {\ro g\over \ro \tilde{h}^{(n)}_i}% & = {P_i\over N}  \pa{{h^{(n)}_i\over y_i} - 1 }\cr
    %&= {P_i\over N} \pa{ \exp\left[\tilde{h}^{(n)}_i -\tilde{y}_i\right] -\mathbf{1}} %\cr &
    \approx {P_i\over N}  \pa{\tilde{h}^{(n)}_i - \tilde{y}_i }.
    \label{eq:lin}
\end{align}
\outNim{
\begin{align}
    {\ro g\over \ro \tilde{h}^{(n)}_i} & = {y_i\over N}\circ  \pa{{h^{(n)}_i\over y_i} - 1 }\cr &= {y_i\over N} \circ \pa{ \exp\left[\tilde{h}^{(n)}_i -\tilde{y}_i\right] -1}  \cr
    & \approx {y_i\over N} \circ  \pa{\tilde{h}^{(n)}_i - \tilde{y}_i }.
    % \label{eq:lin}
\end{align}
}%%%%%
To further simplify this, we define the ``optimal input'' $\ba{h}^{(0)}_i$ as the input that would produce exactly the output vector $y_i$.
In practice, $\ba{h}^{(0)}_i$ can be constructed via activation  maximization \cite{mahendran2016visualizing}. %, starting from a random input.
To be precise
\begin{equation}
    \tilde{h}^{(n)}_i = F[h^{(0)}_i], \quad \tilde{y}_i = F[\ba{h}^{(0)}_i]
    \label{eq:hbar}
\end{equation}
where $F(\cdot)$ summarizes propagation through the network.
Note that $\ba{h}^{(0)}_i$ is not unique for many reasons including nonlinearity of the network, as well as weights not being full-rank.
Let $\ba{h}^{(k)}_i$ denote the raw output of layer $k$ after propagating $\ba{h}^{(0)}_i$ through the network.
% We may use any such $\ba{h}^{(0)}_i$ to expand \eqref{eq:lin} further.
% In particular,
We can always find $\ba{h}^{(0)}_i$ such that the activation patterns of  $\ba{h}^{(k)}_i$ and $\tilde{h}^{(k)}_i$ are similar, meaning $\theta \pa{\ba{h}^{(k)}_i} \sim \theta \pa{ \tilde{h}^{(k)}_i}$. This means that we choose $\ba{h}^{(k)}_i$ such that it uses features similar to $\tilde{h}^{(k)}_i$ in each layer. This ensures that $A^{(k)}$ will be the same for $\ba{h}^{(k)}_i$ and $\tilde{h}^{(k)}_i$ \footnote{This is a reasonable assumption as we are close to convergence and %weights are almost trained.
%Also, $\ba{h}^{(k)}_i$ are not unique and
we can always find an $\ba{h}^{(k)}_i$ which is close enough to $\tilde{h}^{(k)}_i$ so that this is satisfied.}.
Defining $\Delta h^{(k)}_i \equiv \mathrm{diag}\pa{\theta \pa{\tilde{h}^{(k)}_i}} \pa{\tilde{h}^{(k-1)}_i-\ba{h}^{(k-1)}_i}$, we get
\begin{align}
    %\Delta h^{(k)}_i &\equiv \mathrm{diag}\pa{\theta \pa{\tilde{h}^{(k)}_i}} \pa{\tilde{h}^{(n-1)}_i-\ba{h}^{(n-1)}_i}\cr
    \tilde{h}^{(n)}_i - \tilde{y}_i &= {w^{(n)}}^T %\pa{\tilde{h}^{(n-1)}-\ba{h}^{(n-1)}}
    \Delta h^{(n-1)}_i %\cr&
    \approx {A^{(k)}}^T%\pa{\tilde{h}^{(k-1)}-\ba{h}^{(k-1)}},
    \Delta h^{(k-1)}_i, \quad \forall k %{\prod_{m=1}^k b^{(n-m)}}
    \label{eq:expand}
\end{align}
where bias-dependent terms exactly cancel.
\outNim{
\footnote{When $\theta \pa{\ba{h}^{(k)}_i} $ and  $\theta \pa{ \tilde{h}^{(k)}_i}$ are not exactly the same, there will be bias-dependent terms with an $b^{(m)}\pa{A^{(m)}(\tilde{h})-A^{(m)}(\ba{h})}$ factor with $m\geq k$.
But this term is small and we may choose  $\ba{h}^{(k)}_i$ such that this term vanishes.
\outNim{
Thus, expanding to layer $k$, the separation of time-scales states that the largest contribution in $\pa{\tilde{h}^{(n)}_i - \tilde{y}_i}$ comes from
$ {A^{(k)}}^T\pa{\tilde{h}^{(k-1)}-\ba{h}^{(k-1)}}$.}
}.
}
Plugging \eqref{eq:expand} into  \eqref{eq:lin}
\outNim{
we find\footnote{ Note that $\mathrm{diag}\pa{\theta \pa{\tilde{h}^{(k)}_i}}^2 = \mathrm{diag}\pa{\theta \pa{\tilde{h}^{(k)}_i}}.$}
%{\color{red} ye kam tozih bedim khate aval az koja umade}
\begin{align}
    % \tilde{h}^{(n)} &\approx {A^{(1)}}^T h^{(0)}\cr
    %\Rightarrow {\ro g \over \ro h^{(n-1)}_i} &=  {y_i\over N}\circ {w^{(n)}}^T {A^{(1)}}^T\pa{ h^{(0)}_i-\ba{h}^{(0)}_i}  ,
    %\Rightarrow
    {\ro g \over \ro \tilde{h}^{(n)}_i} &\approx %{y_i \over N}\circ {A^{(1)}}^T\pa{ h^{(0)}_i-\ba{h}^{(0)}_i} ,
    {1 \over N} P_{i} {A^{(k)}}^T
    %\pa{ \tilde{h}^{(k-1)}_i-\ba{h}^{(k-1)}_i}
    \Delta h^{(k-1)}_i \quad \forall k,
    \label{eq:dgdhn}
\end{align}
%for any $k$ and again assuming activation patterns are similar, i.e. $\theta \pa{\ba{h}^{(k)}_i} \sim \theta \pa{ \tilde{h}^{(k)}_i}$.
% {\color{red} check this, A's are not the same one is with the bar for h bar}
}
and substituting %\eqref{eq:dgdhn}
in \eqref{eq:dgw}, the weight gradients become
\begin{align}
    {\delta g\over \delta {w^{(k)}}^T}% & =  h^{(k-1)} \pa{{\ro g\over \ro  \tilde{h}^{(n)}}}^T {A^{(k+1)}}^T \cr
    & \approx  {1 \over N}\sum_{i=1}^N h_i^{(k-1)} \Delta {h^{(k-1)}_i}^T %\pa{ h^{(k-1)}_i-\ba{h}^{(k-1)}_i}^T
    {w^{(k)}} K^{(k+1)}_i
    ,\cr
    K^{(k)}_i &\equiv %\pa{ {A^{(m)}}\circ y_i }  {A^{(m)}}^T
    A^{(k)} P_{i}  {A^{(k)}}^T
    \label{eq:dgdw1}.
\end{align}
Note that, while ${h}^{(0)}_N$ is the actual $N$-th input, $\ba{h}^{(0)}_N$ is the best guess for what input would yield output $y_N$ based on information in the $N-1$ previous data points.
Since $\ba{h}^{(k)}_i$ produces exactly the same redefined label $\tilde{y}_i$ for all $i<N$ we have ${A^{(k+1)}}^T\ba{h}^{(k)}_i= {A^{(k+1)}}^T \tilde{h}^{(k)}_i$, %{\color {red} nima man hamchenan ba in moshkel daram hbar nabayd bashe? deltah ham ke tebghe htilde tarif shode}
which results in the first $N-1$ points canceling in \eqref{eq:dgdw1}
\begin{align}
    {\delta g\over \delta {w^{(k)}}^T} &\approx  {1 \over N} h_N^{(k-1)} \Delta {h^{(k-1)}_N}^T %\pa{ h^{(k-1)}_i-\ba{h}^{(k-1)}_i}^T
    {w^{(k)}} K^{(k+1)}_N
    \label{eq:dgdwN}.
\end{align}
which also makes use of the fact that in the relaxation phase the weights are close to their locally optimal value.
Eq. %\eqref{eq:dgdw1} and
\eqref{eq:dgdwN} is a stochastic equation, but the distribution of its solutions can be calculated.

\subsection{Estimating distribution of weights for low layers}
% Since in the relaxation phase gradients are vanishingly small,
%In the relaxation phase, we are near a local minimum and the potential, meaning $|\ro g/ \ro h_i^{(0)} | \ll 1 $.
% adding the $N$-th data point should only result in statistical $1/\sqrt{N}$ fluctuations.
% Eq. \eqref{eq:dgdw1} is a stochastic equation, but the distribution of its solutions can be calculated.
%We will use this assumption to solve the lowest layer.
%{\color{red} Define $\ba{X} = (\ba{h}^{(0)}_1, ..., \ba{h}^{(0)}_N)$. }
% As $w^{(k)}$ varies much faster than $w^{(m)}$ with $m>k$, we can assume that $K_i^{(k+1)}$ in \eqref{eq:dgdw1} is very slowly varying {\color{red} ino nafahmidam che rabti dare?}.
$K_i^{(m)}$ is similar to a projection onto label $y_i$, which is a one-hot vector that is nonzero only for some class $c$.
But different inputs $h_i^{(0)}$ belonging to the same class $c$ may still have very different activation patterns $\theta \pa{\tilde{h}^{(k)}_i}$ in different layers because they may contain different low and high level features.
As a result, $A_i^{(k)}$ can be different for inputs belonging to the same class.
But it is also likely that there are groups in inputs for each class such that within each group the activation patterns are the same (i.e. all inputs in one group use exactly the same low and high level features) and so have exactly the same $A_i^{(k)}$.
For such a group of inputs within a class $c$, $K_i^{(k)}$ will be exactly the same, allowing further simplification of \eqref{eq:dgdw1}.
However, since we do not know a priori what groups of inputs use the same features, even though these groups exist, we lack the information needed to perform such grouping.
So, in the most general case, there does not seem to be any option other than methods like SGD to solve \eqref{eq:dgdw1}.

Despite the above-mentioned problem, in many datasets with refined labels most inputs of the same class share characteristic features.
For instance, in the MNIST dataset of handwritten digits with 10 labels all inputs of the class of number 1 contain vertical lines which are less common in other numbers.
Therefore, the average of $K_i^{(k)}$ over class 1 will be a good approximation of the characteristic $K_i^{(k)}$ of this class, and it will be different from $K_i^{(k)}$ of other classes.
%Such an average would be much less specific to the class if the labels were, say, being even or odd digits.
Thus, if for most inputs within class $c$, we could ignore the variability of the activation pattern $\theta \pa{ \tilde{h}^{(k)}_i}$, we could define a single, averaged $K_i^{(k)}$, denoted by $K_c^{(k)}$, for the whole class $c$. %{\color{red} man gij shodam i be ye noghte barmigarde?  az eq. 10 ba ham morur konim}
%$K_c^{(m)}$ can also be constructed by averaging over inputs belonging to the same class.
%If this simplifying assumption holds we can write
% we have a distinct $K_i^{(m)}$ which we will denote as $K_c^{(m)}$.
% As $w^{(k)}$ varies much faster than $w^{(m)}$ with $m>k$, we can assume that $K_i^{(k+1)}$ in \eqref{eq:dgdw1} is very slowly varying.
%They have an interesting structure.
%$A^{(m)}$ has dimensions $d^{(m-1)} \times C$ and $y_i$ has dimensions $C$.
%Thus $ K_i^{(m)}$, which is $d^{(m-1)}\times d^{(m-1)} $, has a ``bottleneck'' at its center which pulls the $d^{(m-1)}$ dimensional input through $C$ channels and checks if it has nonzero value for channel $y_i$.
%In other words, $K_i^{(m)}$ is a projection operator which only keeps the input channels relevant to the class $c$ {\color{red} inja yekam vazeh tar begim manzur az c kelasi hast ke 1 hast tu vector e y} of label vector $y_i$.
%Thus, for each class $c$ we have a distinct $K_i^{(m)}$ which we will denote as $K_c^{(m)}$.
In \eqref{eq:dgdw1}, since all inputs of the same class %for inputs $i$ whose labels belong to the same class
$c$ have the same $K^{(k)}_c$, %independent of $i$,
we can break the sum over $i$ down to summations over classes.
%Define $X_c$ as all inputs with label class $c$.
\outNim{
Also define $\ba{X}_c = \{\ba{h}^{(0)}_i; i\in c\}$ as the set of $\ba{h}^{(0)}$ for class $c$.
In the first layer, $K_i^{(2)}$ is dragging the output of the first layer through the bottleneck.
%Define $\ba{X} = (\ba{h}^{(0)}_1, ..., \ba{h}^{(0)}_N)$.
We define the ``density matrix''\footnote{ When the inputs $h^{(0)}_i$ are mean zero, the density matrix is just the covariance matrix.} $\rho_c$ of the dataset for each class $X$, and the ``target density matrix'', $\ba{\rho}_c$ as
\begin{align}
     \rho_c(N) \equiv {1\over N}X_cX_c^T &= {1\over N}\sum_{i\in c} h^{(0)}_i  {h^{(0)}_i}^T, \cr
     \ba{\rho}_c(N) \equiv {1\over N} X_c\ba{X}_c^T &= {1\over N}\sum_{i\in c} h^{(0)}_i  {\ba{h}^{(0)}_i}^T\label{eq:rhoc}
\end{align}
}%%%%%
\outNim{
\begin{align}
     \rho(N) \equiv {1\over N}XX^T &= {1\over N}\sum_{i=1}^N h^{(0)}_i  {h^{(0)}_i}^T, \cr
     \ba{\rho}(N) \equiv {1\over N} X\ba{X}^T &= {1\over N}\sum_{i=1}^N h^{(0)}_i  {\ba{h}^{(0)}_i}^T\label{eq:rho}
\end{align}
}%%%%
%where $N_c$ is the number of training samples which belonged to class $c$ in the $N$ data points used so far.
First, define the ``density matrix''\footnote{ When $h^{(k)}_i$ are mean zero, the density matrix is just the covariance matrix.} $\rho_c^{(k)}$ for each class $c$ as%, and the ``target density matrix'', $\ba{\rho}_c^{(k)}$ at step $N$ as
\begin{align}
     \rho_c^{(k)}(N) &\equiv
     {1\over N}\sum_{i\in c} h^{(k)}_i  {h^{(k)}_i}^T. %\cr
     %\ba{\rho}_c^{(k)}(N) &\equiv
     %{1\over N}\sum_{i\in c} h^{(k)}_i  {\ba{h}^{(k)}_i}^T
     \label{eq:rhoc}
\end{align}
%{\color{red} Change: $\ba{\rho}$ is not similar to $\rho$. The point is $2 h(h-\ba{h})^T \sim \Delta \rho(N) $. So we don't need to define $\ba{\rho}$, just have to explicitly calculate $\Delta \rho$ assuming $\ba{h}$ is the optimal solution. }
\outNim{
Note that, while ${h}^{(0)}_N$ is the actual $N$-th input, $\ba{h}^{(0)}_N$ is the best guess for what input would yield output $y_N$ based on information in the $N-1$ previous data points.
Since $\ba{h}^{(k)}_i$ produces exactly the same raw label $\tilde{y}_i$ for all $i<N$ we have $A^{(k+1)}\ba{h}^{(k)}_i= A^{(k+1)}{h}^{(k)}_i$, which results in the first $N-1$ points canceling in \eqref{eq:dgdw1}
\begin{align}
    {\delta g\over \delta {w^{(k)}}^T} &\approx  {1 \over N} h_N^{(k-1)} \Delta {h^{(k-1)}_N}^T %\pa{ h^{(k-1)}_i-\ba{h}^{(k-1)}_i}^T
    {w^{(k)}} K^{(k+1)}_N
    \label{eq:dgdwN}.
\end{align}
}
In \eqref{eq:dgdwN} $\ba{h}^{(k)}_N$ is the guessed input based on the $N-\delta N$ previous inputs and it can be expressed as a linear combination\footnote{Note, $h_i^{(k)}$ may be overcomplete and not be linearly independent, but $\ba{S}$  only needs to map onto a linearly independent subset of them. }
of previous inputs which belonged to the same class $c$ as the last label $y_N$.
%{\color {red} inja $y_N$ set of N previous labels e? ba'd dari migi hbar az N-deltaN be dast miad. Pas label ha ham bayad tebghe hamun input ha tarif shan}.
\outNim{
$$\ba{h}^{(k)}_N = \sum_{i\in c}^{N-\delta N}  h_i^{(k)} \ba{S}_i^T, \qquad \ba{S}\ba{S}^T = I $$
%with $\ba{S}\ba{S}^T = I$,
and consequently $\ba{h}^{(k)}_N {\ba{h}^{(k)}_N}^T = {1\over N} \rho_c^{(k)}$.
}
The last input ${h}^{(k)}_N$, on the other hand, contains new information and cannot be an orthogonal transformation of the previous data.
\outNim{
Thus we write ${h}^{(k)}_N = \sum_{i\in c}^{N-\delta N}  h_i^{(k)} (S+\delta S)_i^T$ where $SS^T=I$, but $S+\delta S$ is not orthogonal.
Since we have freedom in choosing $\ba{h}^{(k)}_N$, we can choose $\ba{S}=S$ and so we have
\begin{align}
    \rho_c^{(k)}(N) =& {N-1\over N} \rho_c^{(k)}(N-1) + {1\over N} h^{(k)}_N {h^{(k)}_N}^T \cr
    =& \rho_c^{(k)}(N-1) + {1\over N}\pa{h^{(k)}_N \Delta {h^{(k)}_N}^T + \Delta h^{(k)}_N {h^{(k)}_N}^T } \cr & +O(\delta S^2)
\end{align}
}%%%%
Using this, we can show explicitly (SI eq.\eqref{eq:hdh-1}) that
% Thus
\begin{equation}
    {1\over N} h^{(k)}_N \Delta {h^{(k)}_N}^T \approx {1\over 2} {\delta \rho_c^{(k)} \over \delta N}\label{eq:hdh}
\end{equation}
And so using \eqref{eq:dgdwN} the SGD equations for the weights become  %for each choice of $\ba{h}_i^{(0)}$ as
\begin{align}
    {{\delta w^{(k)}}\over \delta N} %& = -\eps {\delta g\over \delta {w^{(k)}}^T}%, & {{\delta w^{(1)}}^T\over \delta N} &
    \approx -{\eps\over 2} \sum_{c=1}^C {\delta \rho_c^{(k-1)} \over \delta N} w^{(k)} {K}^{(k+1)}_c\label{eq:dwdN}.
\end{align}
%can be proven explicitly for linear regression (SM)% sec.\ref{sec:reg})
%and since the relaxation phase is also a convex optimization phase, we expect this to be exact, at least for lower layers which are  propagating through fewer layers and are, thus, require less fine-tuning of the layers.
%After a large enough $N$ and assuming that the  training set is uniform\footnote{By uniform we mean that any large sample taken from the dataset has approximately the same covariance matrix.}, we expect $\rho_c^{(0)}$ to represent the statistics of the data.

\begin{figure}[h]
    \centering
    \includegraphics[width=.8\columnwidth%,trim=0 .3cm 0 1cm
    ]{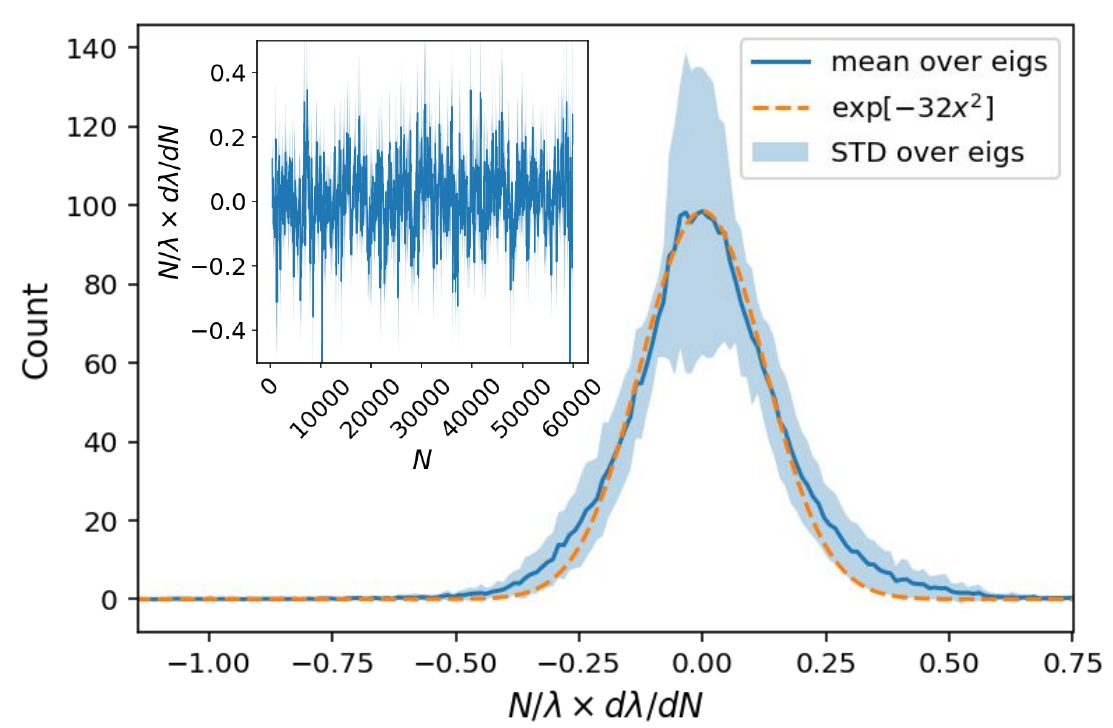}
    \caption{MNIST showing Var$[{\delta \rho \over \delta N}]\propto {\rho^2\over N^2} $. The fluctuations in the eigenvalues $\delta \lambda_\mu /\delta N$ of the covariance matrix takes a random  Gaussian distribution with zero mean and constant variance over added samples $N$ when scaled by $N/\lambda $ (inset).
    Averaging this distribution over all eigenvalues confirms that they all have the same $\lambda^2/N^2$ variance pattern.
    %\vspace{-15pt}
    }
    \label{fig:sigma}
\end{figure}

% \subsection{Dynamics of Relaxation of the Density Matrix}
In the relaxation phase %, the gradient is small and the entries of
$\rho_c^{(k)}$ fluctuates mostly due to statistical fluctuations in the data.
%Thus, fluctuations of $\rho$ must be consistent with standard error, going to zero as $N\to \infty$.
%{\bf Standard Error of Mean of $\rho$ \label{sec:error}}
%We want fluctuations of $\rho$ with addition of data (increasing $N$).
%$\rho = {1\over N}\sum_i^N x_ix_i^T$, where $x_i$ are vectors of observations of the random variable $X$ with $d$ dimensions.
As every input is an independent drawing from the dataset, $\rho_c^{(k)}$ is the sum of $N$ observations $ h^{(k)}_i  {h^{(k)}_i}^T/N$.
% Thus, %$ \mathrm{Var}[\rho] = N \mathrm{Var}[R]$, but
% $ \mathrm{Var}[\delta \rho_c^{(k)}] = \delta N \mathrm{Var}[R]$ as it contains $\delta N$ samples.
% Using the Bienaym\'e formula \cite{loeve1977graduate} and the  fact that $R$ is quadratic in $H =\{h^{(k)}_i\}$, we have $\mathrm{Var}[R] = {\mathrm{Var}[H^2]\over N^2}$.
Since mean and variance of the input do not diverge, the Central Limit Theorem implies that $\rho_c^{(k)}$ will have a multivariate Gaussian distribution and it is straightforward to show that (SI eq.\eqref{eq:rho-fluct-1}) the variance is
% Using Gaussianity $H$ we can directly calculate %and if $H$ is mean zero
\begin{align}
    %\mathrm{Var}[H^2] &= E[H^4] - E[H^2]^2 = 2 \mathrm{Var}[H]^2 = 2 {\rho^{(k)}_c}^2\cr
    %& = 2 \pa{\rho^{(k)}_c - E[H]^2}^2 \cr
    %\Rightarrow
    \mathrm{Var}\left[\delta \rho^{(k)}_c\right] &= 2\delta N {{\rho^{(k)}_c}^2 \over N^2}
    \label{eq:rho-fluct}
\end{align}
Therefore $\delta\rho^{(k)}_c/\delta N$ is also a Gaussian with mean zero and the above variance and we can write
${\delta \rho_c^{(k)} \over \delta N} = \mathcal{N}(0,1){2\over \sqrt{\delta N}} { \rho_c^{(k)} / N}$.
Fig. \ref{fig:sigma} shows the fluctuations of eigenvalues of $\rho^{(0)}$ for MNIST, where the input is broken into $5\times 5$ windows convolved over the images (i.e. input for a convolutional layer).
$\rho^{(0)}$ is 25 dimensional.
When scaled by our prediction of the behavior \eqref{eq:rho-fluct} of the fluctuations, the distribution of the fluctuations of all 25 eigenvalues collapse to a single Gaussian with small error bars, confirming our prediction.

\subsection{Solving the weight distribution}
Putting all the above together, the weight SGD equations become
\begin{align}
    {{\delta w^{(k)}}\over \delta N} & = - \eps(N)\sum_c \rho_c^{(k-1)} {w^{(k)}} K^{(k+1)}_c.\label{eq:dwdN2}
\end{align}
where $\eps(N) \equiv {\eps\mathcal{N}(0,1)\over N \sqrt{\delta N}}$, is a random Gaussian noise with correlation function
% \begin{equation}
    $\bk{\eps(N)\eps(N')}= {\eps^2 \over  N^2 \delta N } \delta(N-N')$.
%\end{equation}
Note that this is a stochastic, ``Langevin equation'' \cite{coffey2004langevin}.
While there are no solutions to it, we can find the probability distribution for different solutions using the corresponding ``Fokker-Planck equation'' \cite{kadanoff2000statistical}.
We do so by first moving the weights $w^{(k)}$ to the right hand side.
\outNim{
$K^{(k)}_c$ is a proportional to a projection matrix onto features relevant to class $c$.
Summing $K_c^{(k)}$ over $c$ yields
\begin{align}
    \sum_c K_c^{(k)} = A^{(k)}\sum_c P_c A^{(k)T} = A^{(k)} A^{(k)T}
\end{align}
% where $P_c$ is a projection matrix onto class $c$.
} %%%%%%
From the separation of time scales, when $w^{(k)}$ in \eqref{eq:dwdN2} has not fully converged, in $A^{(k+1)}$ the $w^{(m)}$ ($m>k$) must be farther from convergence than $w^{(k)}$ and must be still fairly random.
Thus rows of $w^{(m)}$ are uncorrelated, resulting in its transpose being approximately proportional to its pseudo-inverse and so $A^{(k)T} \approx  a_k {\ba{A}^{(k)}}^{-1}$
with $a_k \equiv ||A^{(k)}||^2/d^{(k-1)} $.
As a result, $K_c^{(k)}/a_k$ is a projection matrix onto class $c$. %, propagating through weights of higher layers.
\outNim{
% making it approximately proportional to an orthogonal matrix, and subsequently
\begin{align}
    A^{(m)T} \approx & a_m \ba{A^{(m)}}^{-1},\quad a_m \equiv  {1\over d^{(m-1)}}||A^{(m)}||^2\cr
    \Rightarrow &\sum_c K_c^{(k)} =  A^{(k)} A^{(k)T} \approx a_m P_C \label{eq:sumK}
\end{align}
$P_C$ being the $d^{(m)}$ dimensional identity projection to a $C$ dimensional subspace.
By the same token, $K_c $ is a projection operator onto class $c$, propagating through all the weights of the higher layers.
} %%%%%
Using SVD, we can define a right-pseudo-inverse ${\ba{w}^{(k)}}^{-1}$ such that
\({w^{(k)}}{\ba{w}^{(k)}}^{-1} = I\).
Define the ``class-restricted weights,'' $w_c^{(k)}$.
\outNim{
, as %{\color{red} inam ye tozih bede, ak+1 bara normalization}
\begin{equation}
    w_c^{(k)} \equiv  w^{(k)}{K}_c^{(k+1)},%\quad  \hat{K}_c^{(k)} \equiv {K_c^{(k)}\over a_k} \label{eq:aK}
\end{equation}
Using SVD, we can define a
}%%%%%%%%
% right-pseudo-inverse ${\ba{w}^{(k)}}^{-1}$ such that \({w^{(k)}}{\ba{w}^{(k)}}^{-1} = I\).
It follows that
\begin{align}
    w_c^{(k)} &\equiv   w^{(k)}{{K}_c^{(k+1)}\over a_k}, &
    {\ba{w}_{c}^{(k)}}^{-1} &\approx {{K}_c^{(k+1)} \over a_k} {\ba{w}^{(k)}}^{-1},
\end{align}
which satisfy $w_{c'}^{(k)} {\ba{w}_c^{(k)}}^{-1} \approx  \delta_{cc'} I $.
Multiplying both sides of \eqref{eq:dwdN2} by ${\ba{w}_c^{(k)}}^{-1}$, we get
\begin{align}
    {{\delta w^{(k)}}\over \delta N} {\ba{w}_c^{(k)}}^{-1} %& = -\eps (N)\sum_{c'} \rho_{c'}^{(k-1)} {w_{c'}^{(k)}}{\ba{w}_c^{(k)}}^{-1} \cr
    & \approx -\eps (N) \rho_c^{(k-1)}.
    \label{eq:dwdN3}
\end{align}
% Note that we still have the full $\delta w^{(k)}/ \delta N $ on the left hand-side. But s
Since $\delta w^{(k)} \gg \delta w^{(m)}  $ for $m>k$, we can write\footnote{
Also, note that because of the $K^{(k+1)}_c$ on the r.h.s. of \eqref{eq:dwdN2} dimensions that do not get mapped through $K^{(k+1)}_c$ also do not appear in the SGD equations.}
\begin{align}
    {\delta {w^{(k)}}\over \delta N} & = \sum_c {\delta {w_c^{(k)}}\over \delta N} \approx \sum_c {\delta {w^{(k)}} \over \delta N} {{K}_c^{(k+1)}\over a_k},
    \label{eq:dwK}\cr
    {{\delta w^{(k)}}\over \delta N} {\ba{w}_c^{(k)}}^{-1} &\approx  {{\delta w_c^{(k)}}\over \delta N} {\ba{w}_c^{(k)}}^{-1} \equiv  {\delta \log_R { w_c^{(k)}}\over \delta N}
\end{align}
where the ``right logarithm,'' is formally defined using the pseudo-inverse for the largest non-degenerate submatrix such that\footnote{
Note that $\log_R  w_c^{(k)} $ is a $d^{(k-1)}\times d^{(k-1)}$ dimensional matrix, but the its largest non-degenerate submatrix is one dimensional, because of the projection via $K_c^{(k+1)}$. }
$\delta  \log_R { w_c^{(m)}} = \pa{\delta w^{(m)}} {\ba{w}_c^{(m)}}^{-1}$.
This suggests that \eqref{eq:dwdN2} could be solved for the class-restricted weights separately, and that for each class we have a stochastic equation given by
\begin{align}
    {\delta \log_R { w_c^{(k)}} \over \delta N} & = - \eps(N) \rho_c^{(k-1)} %\equiv -\eta_c(N)\cr
    % \bk{\eta_c(N)\eta_c^T(N')} & =  {\eps^2 \over  N^2 \delta N }{\rho_c^{(k-1)}}^2  \delta(N-N')
\end{align}
This is a Langevin equation with a multivariate noise distributed according to $\rho_c^{(k-1)}$ whose magnitude decreases  with increasing $N$.
%This is simply a multi-variate Brownian motion, but instead of expanding with increasing steps $N$, its width actually decreases with $N$.
The corresponding Fokker-Planck equation has solution (for simplicity $v\equiv \log_R  w_c^{(k)}$ )
\begin{equation}
    \Pi(v) = Z^{-1} \exp\left[-{1\over \bk{ \eps(N)^2} } v^T{\rho_c^{(k-1)}}^{-2}v \right]
    \label{eq:FP}
\end{equation}
which is a multivariate Gaussian distribution with spread $\sigma = \rho_c^{(k-1)}/\sqrt{\bk{ \eps(N)^2}}$ proportional to the class density matrix.
%The distribution $\Pi(v)$ is not uniform over the space of $v$.
It spreads along directions of eigenvectors $\psi_\mu^c$ of $\rho_c^{(k-1)}$, spreading wider along eigenvectors with larger eigenvalues.
\outNim{
Just as in diffusion, $v$ is wider in the directions of the eigenvectors $\psi_\mu^c$ with larger eigenvalues $\lambda_\mu^c$.
Specifically, the width of the distribution of $v$ along $\psi_\mu^c$ satisfies
\begin{equation}
    \sigma_v\cdot \psi_\mu^c = \eps_0\sqrt{{\lambda_\mu^c\over N}} \psi_\mu^c. \label{eq:sigFP}
\end{equation}
This means that the distribution $v$ has larger component along larger eigenvalue eigenvectors of $\rho_c$, similar to PCA {\color {red} ino mishe ghavi tar neshun dad. Axi?}.
We can also find the approximate of $w^{(1)}_c$ using a similar reasoning, as we discuss below. }
This means that for each class, just as in PCA, there are a subset of weights $w^{(k)}_c$ which are likely to be along principal components (PC) of $h^{(k)}_i$.
Since $w K = wSS^T K$ for any orthogonal $S$, we are free to choose the basis of the output of the weights for each layer (i.e. the choice of basis is part of the design of the network architecture).
We will therefore choose distinct rows of $w^{(k)}$ to be dedicated to each class, trivially satisfying $w^{(k)}_c {w^{(k)}}^T_{c'} \propto \delta_{cc'}$.
Then, for each class the most likely outcomes of SGD will be that rows of $w^{(k)}_c$ are PC of $h^{(k)}_i$.
If a PC of one class is highly correlated with a PC of another class, we can keep one of them.
The biases $b^{(k)}$ canceled in \eqref{eq:expand} and so in the late stages of training they don't seem to play a significant role.
We may choose $b^{(k)}=0$.
Thus, we can construct pretrained neural networks with weights found using the class-based PCA described here and with biases set to zero.
We will call such a network a ``Density Matrix Network'' (DMN).

\outNim{
In the end, note that the assumption that the projections $K_i^{(k)}$ can be averaged over one class relied on the assumption that different elements in the same class use similar features.
While this may be usually true for low-level features, it is more likely to fail for high-level features.
For lower layers (smaller $k$) the similarity of features in lower layers may mask this problem and result in similar $K_i^{(k)}$ for most class inputs, but in higher layers the problem should become more evident.
Thus, we expect this method to perform worse in higher layers.
} %%%%%%

\begin{figure*}[ht]
    \centering
    \includegraphics[width=\textwidth,trim=0 2cm 0 2cm]{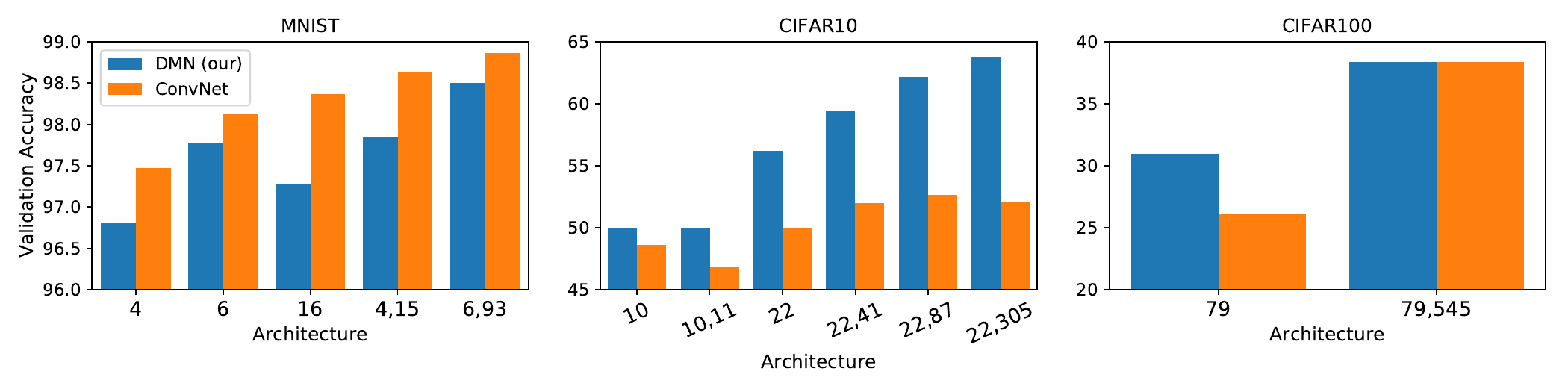}
    \caption{{\bf DMN with one or two layers versus ConvNet:}
    The architectures for the DMN and the ConvNet are chosen to be exactly the same with the same number of filter and layers.
    The y-axis shows the validation accuracy (percentage) and
    the x-axis labels show the number of filter (``4'' means a single  convolutional layer with 4 filters; ``4,15'' means two layers, first with 4 and second with 15 filters).
    On MNIST (left) DMN (blue) performs slightly (about 1 percent accuracy) worse than ConvNet (orange) almost in all tested cases.
    Yet, the mere fact that  a pre-computed network is performing comparable to to a trained network is noteworthy.
    On CIFAR10 (center), DMNs consistenly outperform the ConvNets, with a very impressive margin of between $5-10\%$ in the two layer setting.
    On CIFAR100 (right), the single layer DMN outperforms ConvNet, while in two layers the performance is equal.
    All experiments have 1 dense classification layer with softmax activation.
    % ``d'' stands for DMN, and ``c'' for ConvNet.
    % ``d4, d15'' means two layers of DMN, first with 4 filters and second with 15.
    Each DMN and ConvNet layer is ReLU activated and is followed by a $2\times2$ maxpooling layer.
    All filters in DMN and ConvNet have $3\times3$ receptive fields.
    }
    \label{fig:results}
\end{figure*}
\outNim{
\begin{figure*}[ht]
    \centering
    \includegraphics[width=.32\textwidth,trim=1cm 2cm 1cm 4cm ]{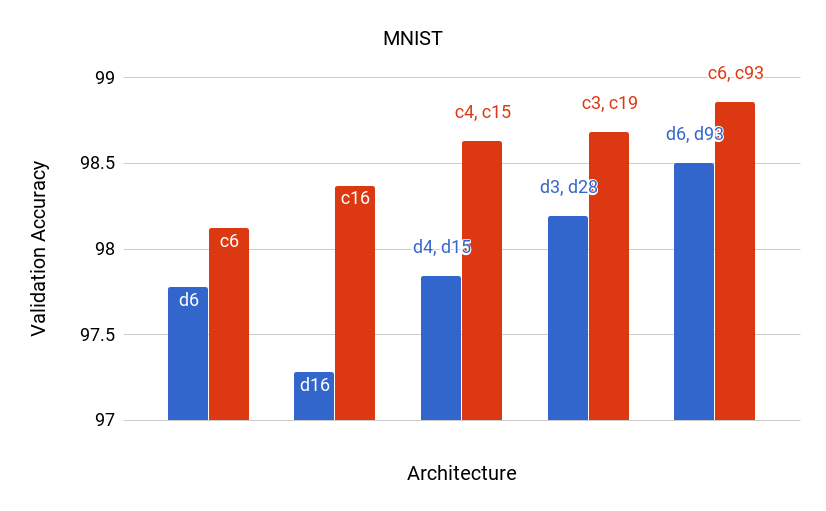} \includegraphics[width=.33\textwidth,trim=1cm 2cm 0cm 4cm ]{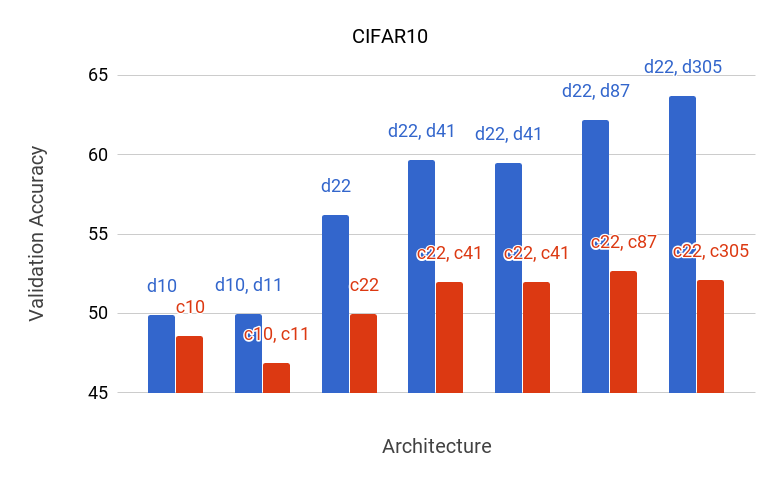} \includegraphics[width=.31\textwidth,trim=1cm 2cm 2.5cm 4cm ]{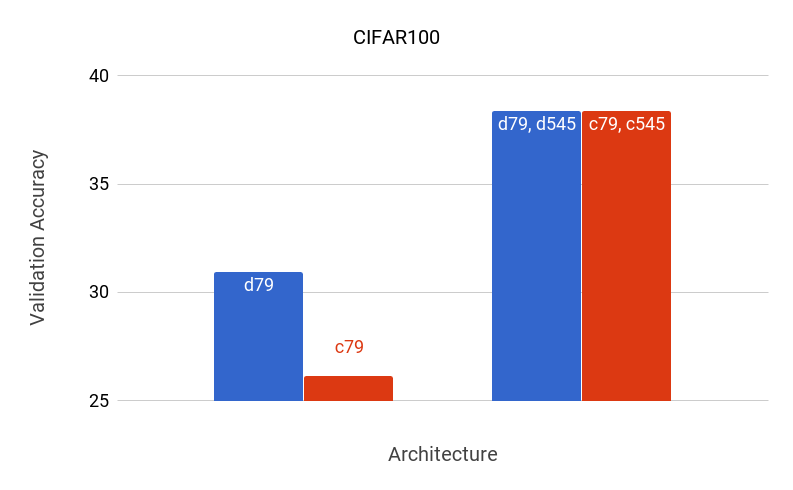}
    \caption{{\bf DMN with one or two layers versus ConvNet:}
    As we see, on MNIST (left) DMN (blue) performs slightly (about 1 percent accuracy) worse than ConvNet (red) almost in all tested cases.
    Still, the mere fact that it is so close in performance to a trained network is noteworthy.
    On CIFAR10 (center), DMNs consistenly outperform the ConvNets, with a very impressive margin of between $5-10\%$ in the two layer setting.
    On CIFAR100 (right), the single layer DMN outperforms ConvNet, while in two layers the performance is equal.
    All experiments have 1 dense classification layer with softmax activation. ``d'' stands for DMN, and ``c'' for ConvNet.
    ``d4, d15'' means two layers of DMN, first with 4 filters and second with 15.
    Each DMN and ConvNet layer is ReLU activated and is followed by a $2\times2$ maxpooling layer.
    All filters in DMN and ConvNet have $3\times3$ receptive fields.
    }
    \label{fig:results0}
\end{figure*}
} %%%
\subsection{Time Complexity}
Training with SGD using $N$ mini-batches of data requires calculating the gradients \eqref{eq:dgw}, which require matrix product of the weights of layers.
Thus the complexity of SGD is
\begin{align}
    T = N\sum_{k=0}^{n-1} O\pa{\prod_{m=k}^n d^{(m)}} = N O\pa{\prod_{m=0}^n d^{(m)}}
\end{align}
If we replace the lowest layer by a DMN, we remove the $d^{(0)}$ factor, but get an additional term for computing the DMN and another for propagating the data through the DMN.
Constructing the DMN requires PCA of all $\rho_c^{(0)}$, or equivalently SVD of $X$, which has complexity $O\pa{N{d^{(0)}}^2}$ (assuming $N > d^{(0)}$).
Propagating the input $X$ through the DMN is $O\pa{N d^{(0)} d^{(1)}}$.
Thus the complexity of replacing a layer with DMN and training the rest of the network via SGD is
\begin{align}
    T =& N \left[O\pa{\prod_{m=1}^n d^{(m)}} + O\pa{d^{(0)}d^{(1)}} + O\pa{{d^{(0)}}^2} \right]
\end{align}
Thus, if $d^{(1)} \sim d^{(0)}$ and $\prod_{m=2}^n d^{(m)} > d^{(0)}$ using DMN will certainly result in a reduction in the time complexity of training.
In particular, in very deep networks it is more likely to have $\prod_{m=2}^n d^{(m)} \gg d^{(0)}$ and thus using DMN can boost the training quite significantly.

\section{Simulation and discussion}
% \section{Conclusion}
%We have shown that, under certain assumptions of uniform sampling in the training dataset, the weights of low-lying layers in deep neural networks converge to a subset of the principal components of the training data.
%The PCA is done in a supervised fashion, using input data relevant to each output class separately.
Training deep neural networks (DNN) requires large amounts of training data and is very computationally expensive.
Part of the reason for this is a lack a good mathematical understanding of solutions a DNN converges into.
%Analyzing the mathematical properties of the stochastic gradient descent (SGD) equations used for training,
%We have shown that stochastic gradient descent (SGD) may often perform a class-based PCA.
%In conclusion,
We have shown that it can be possible to {\em derive} optimal weights for low-lying layers of a DNN directly from data using a class-based PCA.
This relied deeply on
%Our calculations were deeply linked with the existence of
a ``separation of time scales'' which occurs in the highest performing setups for DNNs.
We found that when the activation function is similar to ReLU (unbounded from above and asymptotically linear) and the weights are initialized with a high enough variance (allowing the weights to acquire singular values larger than one) lower layers (closer to input) evolve much faster than layers above them.
This effectively decouples the SGD dynamics of lower layer from layers above and so SGD will be able to find optimal weights layer-by-layer, starting from closest to the input data.
The solutions found by SGD in these cases will be a class-based PCA.

Based on this finding, we can construct pretrained layers using class-based PCA for the weights and setting biases to zero.
We will call such a pretrained layer a ``Density Matrix Network'' (DMN).
The class-based PCA consists of calculating the eigenvectors of the covariance matrix $\rho_c$ of input for each class $c$ and keeping the eigenvectors with  the largest eigenvalues.
The number of eigenvectors to keep depends on how much of the variance of the input data we wish to keep.
Additionally, $\rho_c$ converges very quickly and using only a fraction of the training data used for SGD will still yield DMN performance comparable to SGD in many datasets.
We have implemented DMN and tested it on three image data sets.
%, MNIST, CIFAR10, CIFAR100 (Fig. \ref{fig:results}).
The tested DMNs perform on par, or far superior to conventional networks with the same architecture.
DMN is within one percent of the performance of a convolutional layers (ConvNet) on MNIST, and significantly superior to ConvNets on CIFAR10.
A single layer DMN also performs much better than a similar ConvNet on CIFAR100, and a two layer DMN is equally good as a 2 layer ConvNet.
The results are summarized in Fig. \ref{fig:results}.
Most of our tested DMNs were trained using a fraction (between 20-70\%) of the training data used for the ConvNet.  %Notice that it takes much less time and computation to calculate weights in a DMN as repeated backpropagation is not needed and a fraction of the training data may be sufficient {\color {red} bebin inja raje be time harf mizanim bayad moghayese run time biarim. Ru hava nemishe :D}.

Note that in these tests DMNs are constructed as convolutional layers\footnote{This is related to the Karhunen-loeve  transform of images, where one breaks an image down into blocks and PCA is performed on the block images. }.
We only have to note that overlapping receptive fields will result in spurious PCs and we need to disentangle the outputs to get rid of them.
One simple way to do so is to use a maxpooling layer right after a DMN.
We return to this below.
The architectures we used for our experiments consist of one or two  convolutional layers with ReLU activation functions, each followed by a maxpooling layer and ending with a classification layer with softmax activation function.
We run the experiments once with DMNs for the convolutional layers and once with regular ConvNet (More experiments with hybrids of DMN and ConvNet, as well Batch Normalization \cite{ioffe2015batch} are presented in the SI).

The supervised PCA, which emerged from SGD, may be related to the ``information bottleneck'' \cite{tishby2000information}, which we will investigate in the future.

\outNim{
\begin{acknowledgments}
-- text of acknowledgments here, including grant info --

\end{acknowledgments}
}

% \bibliography{2018/mybib}
% \bibliographystyle{abbrv} % plain unsrt

\end{article}

% \newpage

% \newpage
% \vspace{5cm}
% \newpage

\renewcommand{\thesection}{SI \arabic{section}}
\setcounter{page}{1}
\setcounter{figure}{0}

%\appendix

%%%%%%%%% TITLE - PLEASE UPDATE
% \renewcommand\thesection{\arabic{section}}
% \renewcommand\thesubsection{\thesection.\arabic{subsection}}

% \makeatletter
% \def\@seccntformat#1{\@ifundefined{#1@cntformat}%
%   {\csname the#1\endcsname\space}%    default
%   {\csname #1@cntformat\endcsname}}%  enable individual control
% \newcommand\section@cntformat{\thesection.\space}       % section-level
% \newcommand\subsection@cntformat{\thesubsection.\space} % subsection-level
% \makeatother
\medskip

{\Large\noindent\bf
Supplemental Information:
\title{Separation of time scales and direct computation of weights in deep neural networks}
}
\bigskip
\\
{\small
Nima Dehmamy$^*$,
Neda Rohani$^\dag$
\and
Aggelos Katsaggelos$^\dag$
}
\medskip
\\
{\small\it
{\noindent$*$
CCNR, Northeastern University, Boston 02115 MA, USA}
\\
{\noindent$\dag$
IVPL, Northwestern University, Evanston, 60208 IL, USA}
}

\bigskip

\section{Smoothing of energy landscape}
\begin{figure*}[h]
    \centering
    \includegraphics[width=.5\columnwidth]{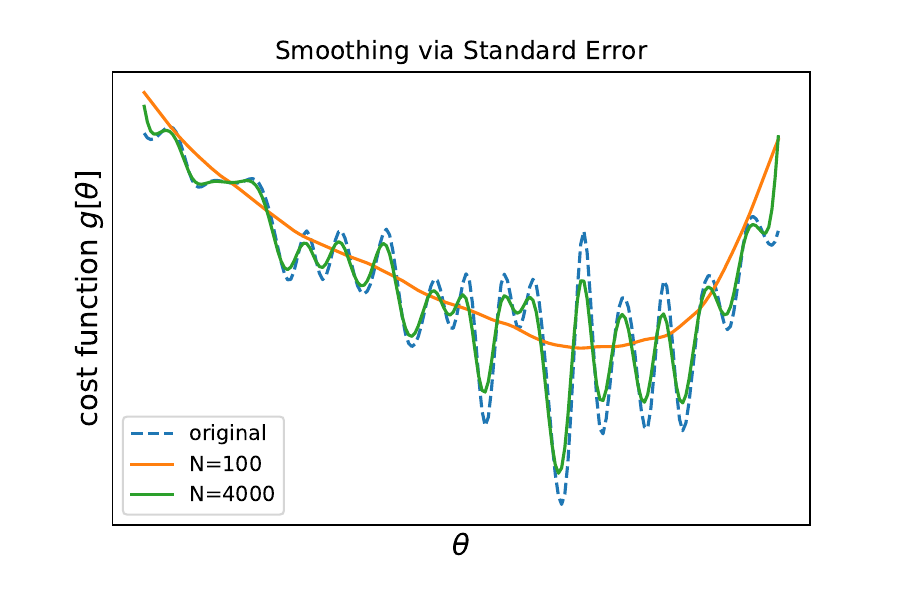}
    \caption{ {\bf Schematic of the smoothing effect of standard error on the landscape of the cost function $g[\theta]$. }
    At step $N$ the standard error in mean of parameters $\theta$ is $\ba{\sigma} = \sigma_\theta/\sqrt{N}$.
    Thus, stochastic fluctuations can move the $\theta$ by a normal distribution of width $\ba{\sigma}$.
    The effective energy landscape is then the convolution of the standard error distribution with the original $g[\theta]$ (dashed blue curve), resulting in a smoothing of the landscape at low $N$ (orange, $N=100$).
    At high $N$, the standard error is negligible and the original $g[\theta]$ is recovered (green $N=4000$).
    %{\color{red} what is g? the formulation also is W one dimension? range of w and g[w] on plot???}
    }
    \label{fig:stderror-schema}
\end{figure*}

\section{Singular Values of weights during SGD}

To quantify the effect of $A^{(k+1)}$, we must examine its singular values and find out how they evolve with training steps, i.e. increasing $N$. % {\color{red} bebin tu har epoch hame point ha barresi mishan yani N sabete shayad step begi behtare. Chon jahaye ghabl ham az step estefade shode o ba epoch ye mani nemide}.
Define $\tilde{w}^{(k)} \equiv \mathrm{diag} \pa{ \theta \pa{ \tilde{h}^{(k-1)}} } w^{(k)} $.
At step zero, weights and biases of all layers are random and so we can assume that half of the entries of $\theta \pa{ \tilde{h}^{(k-1)}}$ will be zero, the other half one.
Thus, in $\tilde{w}^{(k)}$ half of the rows of $w^{(k)}$ will be replaced by zeros.
%{\color{red}being nonzero we can remove this chon w ha ke sefr nistan wtilde sefr mishe, injuri bad khunde mishe}.
Consider the symmetric positive semi-definite $d^{(k)}\times d^{(k)}$ matrix $M^{(k)}\equiv \tilde{w}^{(k)T} \tilde{w}^{(k)}$.
The eigenvalues of $M^{(k)}$ are squares of singular values (SV) of $\tilde{w}^{(k)}$.
The $w^{(k)}$ are generally initialized randomly with zero mean.
As columns of $w^{(k)}$ are uncorrelated at the start, $M$ will be approximately diagonal and all diagonal entries will be similar because %{\color{red} in tikke o Eq. ro tozih bede}
\begin{align}
    M_{ab}^{(k)} &= \sum_c \tilde{w}^{(k)}_{ca}\tilde{w}^{(k)}_{cb} \approx  {d^{(k-1)} \sigma^2_{(k)a}\over 2} \delta_{ab} \label{eq:Mk},%\cr
    % \sigma_{(k)i}^2 &= {1\over d^{(k-1)}} \sum_j \pa{\tilde{w}_{ji}^{(k)}}^2
\end{align}
where $\sigma_{(k)a}^2 = \sum_b \pa{w_{ba}^{(k)}}^2/ d^{(k-1)}$ is the variance of the weights over the input dimension and the $d^{(k-1)}/2$ is because of $\theta \pa{ \tilde{h}^{(k-1)}}$ eliminating half of the rows.
The variance of all rows is chosen to be the same $\sigma_{(k)a}^2=\sigma_{(k)}^2 $ initially.
Eq. \eqref{eq:Mk} implies that most eigenvalues of $M^{(k)}$ are initially close to $d^{(k-1)}\sigma_{(k)}^2/2$.
Now consider $B^{(k)} \equiv {A^{(k)}}^TA^{(k)}$ whose eigenvalues are squares of SV of $A^{(k)}$.
Using \eqref{eq:Mk} we can progressively simplify $B^{(k)}$ and get
%{\color{red} in tikke o Eq. ro tozih bede}
\begin{align}
    B^{(k)} & \approx {d^{(k-1)} \sigma^2_{(k)} \over 2}  {A^{(k+1)}}^TA^{(k+1)} \approx \prod_{m=k}^n {d^{(m-1)} \sigma^2_{(m)} \over 2} I
    \label{eq:Bk}
\end{align}
where
%$\ba{\sigma}^2_{(m)} = \sum_a \sigma^2_{(m)a}/d^{(k)}$ is the variance averaged over rows and
$I$ is the $C\times C$ identity matrix.
Experiments have shown \cite{he2015delving} that with ReLU, the Glorot initialization \cite{glorot2010understanding} (a variant of the Xavier method) which sets $\sigma_{(k)i}^2 = 2/ d^{(k-1)}$ yields better performance than other commonly used initialization methods.
Glorot initialization was designed specifically to set the maximum SV of all $w^{(k)}$ around one to avoid explosion of gradients, while choosing a smaller initialization makes the gradients too small and leads to worse results.
However, this initialization does not guarantee that the maximum SV will {\em remain} below one.
In fact, we argue that the reason this choice works better than a smaller initialization is precisely because the SV become larger than one early in SGD.

\begin{figure*}[h]
    \centering
    \includegraphics[width=.5\textwidth]{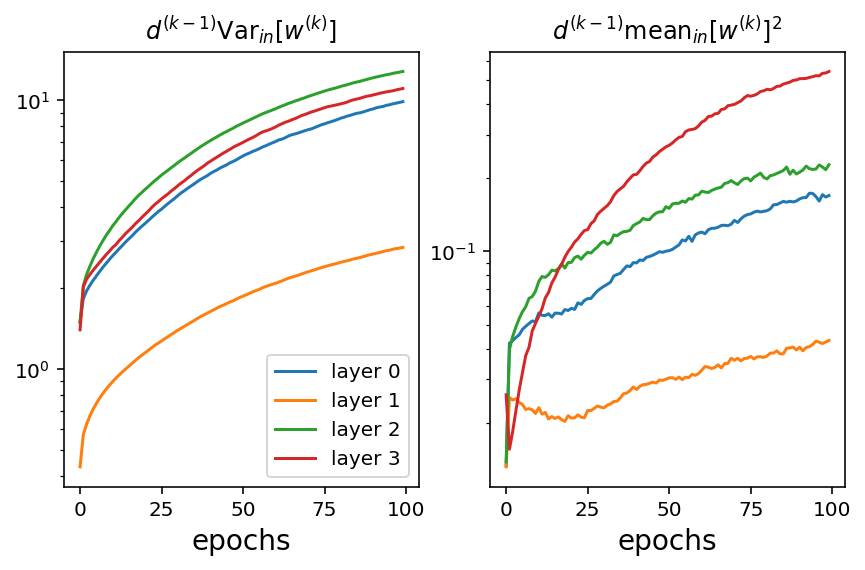}
    \caption{Evolution of mean (right) and variance (left) over input dimensions of weights.
    The square of mean over the input $\pa{\sum_i w_{ij}^{(k)}}^2/d^{(k-1)}$
    }
    \label{fig:meanvar}
\end{figure*}

\subsection{Magnitude of $h^{(k-1)}$}
One may argue that $h^{(k)}$ contains the product of the weights of the first $k$ layers and therefore the gradients of $w^{(k)}$ for all $k$ are of the same magnitude.
But it is easy to see that $h^{(k)}$ cannot remain $ h^{(k)} \sim \prod_{m=1}^k w^{(m)} h^{(0)}$.
To see this, we examine the change in $\delta h^{(k)}$ after processing a $\delta N$ minibatch during SGD.
Using \eqref{eq:SGD}--\eqref{eq:dgw}, we have
\begin{align}
    {\delta \tilde{h}^{(k)} \over \delta N}
    =& -\eps \left[ \pa{1+ |h^{(k-1)}|^2} A^{(k+1)} {\ro g \over \ro \tilde{h}^{(n)}}
    + {w^{(k)}}^T {\delta h^{(k-1)} \over \delta N} \right]\cr
    = & -\eps \sum_{m=1}^{k-1} \pa{1+ |h^{(k-m)}|^2} W^{(m)T}W^{(m)}  A^{(k+1)} {\ro g \over \ro \tilde{h}^{(n)}}\cr W^{(m)}\equiv &\prod_{p=k-m+2}^k \mathrm{diag} \pa{ \theta \pa{ \tilde{h}^{(p-1)}} } w^{(p)}
    \label{eq:dhk-1}
\end{align}

\begin{figure*}%[h]
    \centering
    \includegraphics[width = .5\columnwidth]{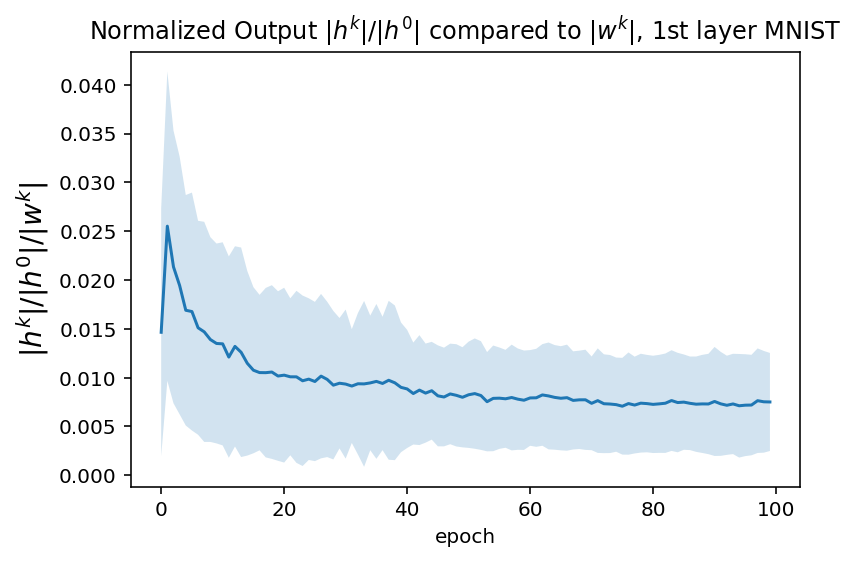}
    \caption{Average of output of the first layer of a ConvNet with 32 filters and ReLU divided by norm of $3\times 3$ filter weights, divided by norm of input image in each $3\times3$ block.
    The plot shows this ratio for different training steps.
    The solid line is mean over all filters and the shaded area is one sigma above and below the mean.
    A small amount of random noise was added to input image to make norms nonzero everywhere, but that did not change the outcome.
    As we see, the output is much smaller than the magnitude of the weights, which suggests that the bias and ReLU diluted the effect of the weights significantly.
    }
    \label{fig:norm}
\end{figure*}

Fig. \ref{fig:norm} shows the magnitude of $h^{(1)} $ divided by the magnitude of the weights $w^{(1)}$ and the input $h^{(0)}$ in a network consisting of a ConvNet with $3\times 3$ kernels followed by a dense classification layer being trained on MNIST.
As we see the maginitude of the output does not become comparable to $\lt{w^{(1)}} \lt{h^{(0)}}$ which itself is larger than $\lt{w^{(1)} h^{(0)}}$ and is, in fact, orders of magnitude smaller.

\section{Distribution of eigenvalues of a random matrix with positive entries}
Recall that the characteristic equation has the form $\lambda^d - \mathrm{Tr}[M]\lambda^{d-1} +...+ \textrm{det}[M] = 0 $, so the sum of the eigenvalues is the trace $t$, which for a random matrix is also the mean of each row.
Wigner's semicircle law \cite{abramowitz1988handbook} states that the eigenvalues are distributed symmetrically around zero with radius $2\sqrt{t}$.
Since the trace is invariant under unitary transformations, it is the sum of the eigenvalues and as the semi-circle eigenvalues are symmetric around zero they all cancel, leaving only the highest eigenvalue.
Thus the largest eigenvalue is the trace.

\section{ Relation of added data and changes in the density matrix}

In \eqref{eq:dgdwN} $\ba{h}^{(k)}_N$ is the guessed input based on the $N-\delta N$ previous inputs and it can be expressed as a linear combination\footnote{Note, $h_i^{(k)}$ may be overcomplete and not be linearly independent, but $\ba{S}$  only needs to map onto a linearly independent subset of them. }
of previous inputs which belonged to the same class $c$ as  $y_N$
$$\ba{h}^{(k)}_N = \sum_{i\in c}^{N-\delta N}  h_i^{(k)} \ba{S}_i^T, \qquad \ba{S}\ba{S}^T = I $$
%with $\ba{S}\ba{S}^T = I$,
and consequently $\ba{h}^{(k)}_N {\ba{h}^{(k)}_N}^T = {1\over N} \rho_c^{(k)}$.
The last input ${h}^{(k)}_N$, on the other hand, contains new information and it cannot be an orthogonal transformation on them.
Thus we write ${h}^{(k)}_N = \sum_{i\in c}^{N-\delta N}  h_i^{(k)} (S+\delta S)_i^T$ where $SS^T=I$, but $S+\delta S$ is not orthogonal.
Since we have freedom in choosing $\ba{h}^{(k)}_N$, we can choose $\ba{S}=S$ and so we have
\begin{align}
    \rho_c^{(k)}(N) =& {N-1\over N} \rho_c^{(k)}(N-1) + {1\over N} h^{(k)}_N {h^{(k)}_N}^T \cr
    =& \rho_c^{(k)}(N-1) + {1\over N}\pa{h^{(k)}_N \Delta {h^{(k)}_N}^T + \Delta h^{(k)}_N {h^{(k)}_N}^T } \cr & +O(\delta S^2)
\end{align}
Thus
\begin{equation}
    {1\over N} h^{(k)}_N \Delta {h^{(k)}_N}^T \approx {1\over 2} {\delta \rho_c^{(k)} \over \delta N}\label{eq:hdh-1}
\end{equation}

\subsection{Dynamics of Relaxation of the Density Matrix}
\begin{figure*}[h]
    \centering
    \includegraphics[width=.7\columnwidth%,trim=0 .3cm 0 1cm
    ]{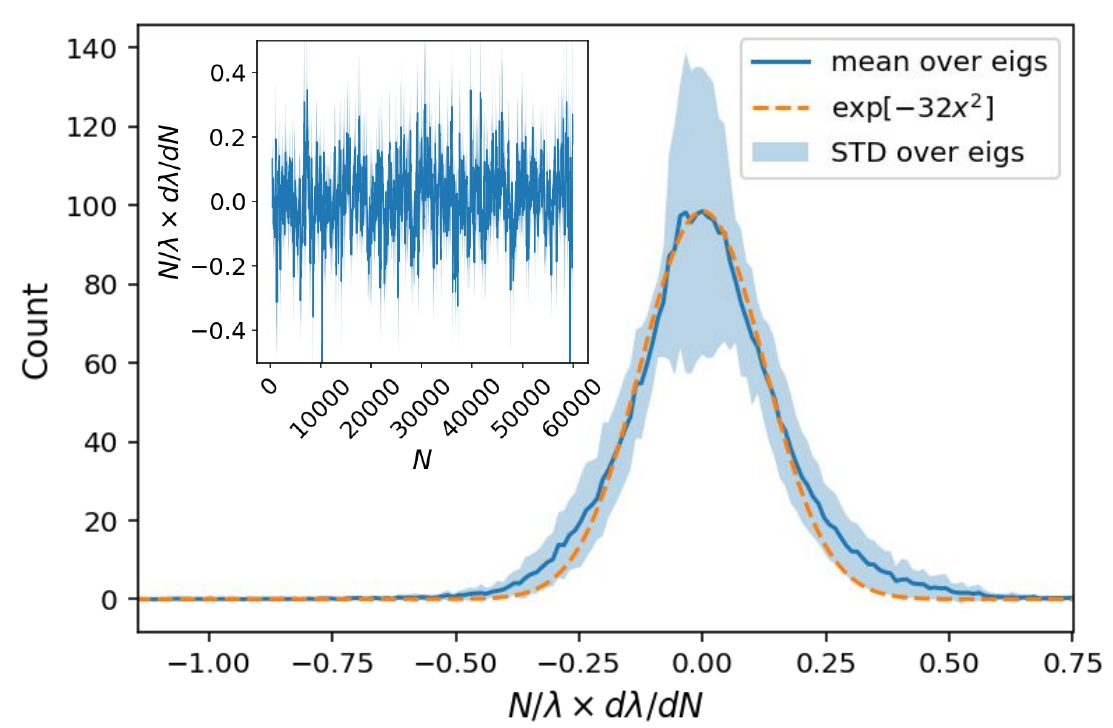}
    \caption{MNIST showing Var$[{\delta \rho \over \delta N}]\propto {\rho^2\over N^2} $. The fluctuations in the eigenvalues $\delta \lambda_\mu /\delta N$ of the covariance matrix takes a random  Gaussian distribution with zero mean and constant variance over added samples $N$ when scaled by $N/\lambda $ (inset).
    Averaging this distribution over all eigenvalues confirms that they all have the same $\lambda^2/N^2$ variance pattern.
    %\vspace{-15pt}
    }
    \label{fig:sigma-1}
\end{figure*}

In the relaxation phase %, the gradient is small and the entries of
$\rho_c^{(k)}$ fluctuates mostly due to statistical fluctuations in the data.
%Thus, fluctuations of $\rho$ must be consistent with standard error, going to zero as $N\to \infty$.
%{\bf Standard Error of Mean of $\rho$ \label{sec:error}}
%We want fluctuations of $\rho$ with addition of data (increasing $N$).
%$\rho = {1\over N}\sum_i^N x_ix_i^T$, where $x_i$ are vectors of observations of the random variable $X$ with $d$ dimensions.
As every input is an independent drawing from the dataset, $\rho_c^{(k)}$ is the sum of $N$ observations $r_i = h^{(k)}_i  {h^{(k)}_i}^T/N$ of a random variable $R$. Thus, %$ \mathrm{Var}[\rho] = N \mathrm{Var}[R]$, but
$ \mathrm{Var}[\delta \rho_c^{(k)}] = \delta N \mathrm{Var}[R]$ as it contains $\delta N$ samples.
Using the Bienaym\'e formula \cite{loeve1977graduate} and the  fact that $R$ is quadratic in $H =\{h^{(k)}_i\}$, we have $\mathrm{Var}[R] = {\mathrm{Var}[H^2]\over N^2}$.
Since mean and variance of the input do not diverge, Central Limit Theorem implies that $\rho_c^{(k)}$ will have a multivariate Gaussian distribution.
Using Gaussianity $H$ we can directly calculate %and if $H$ is mean zero
\begin{align}
    \mathrm{Var}[H^2] &= E[H^4] - E[H^2]^2 = 2 \mathrm{Var}[H]^2 = 2 {\rho^{(k)}_c}^2\cr
    %& = 2 \pa{\rho^{(k)}_c - E[H]^2}^2 \cr
    \Rightarrow \mathrm{Var}\left[\delta \rho^{(k)}_c\right] &= 2\delta N {{\rho^{(k)}_c}^2 \over N^2}
    \label{eq:rho-fluct-1}
\end{align}
Therefore $\delta\rho^{(k)}_c/\delta N$ is also a Gaussian with mean zero and the above variance and we can write
${\delta \rho_c^{(k)} \over \delta N} = \mathcal{N}(0,1){2\over \sqrt{\delta N}} { \rho_c^{(k)} / N}$.
Fig. \ref{fig:sigma-1} shows the fluctuations of eigenvalues of $\rho^{(0)}$ for MNIST, where the input is broken into $5\times 5$ windows convolved over the images (i.e. input for a convolutional layer).
$\rho^{(0)}$ is 25 dimensional.
When scaled by our prediction of the behavior \eqref{eq:rho-fluct} of the fluctuations, the distribution of the fluctuations of all 25 eigenvalues collapse to a single Gaussian with small error bars, confirming our prediction.

\section{Convergence of layers}
\begin{figure*}[ht]
    \centering
    \includegraphics[width = .85\columnwidth]{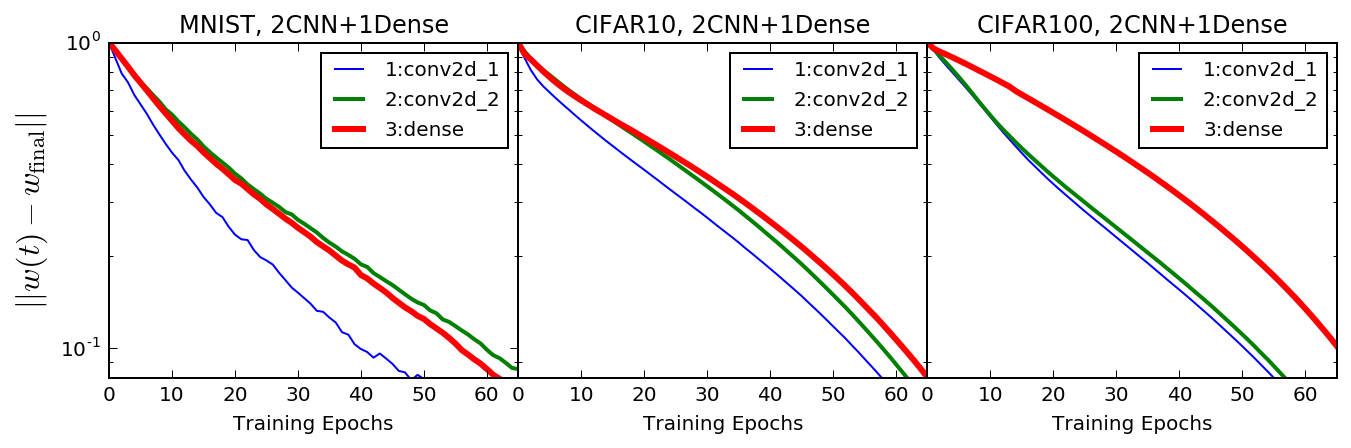}%\includegraphics[width = .3\columnwidth]{figs-paper/energies.png}
    \caption{%{\bf Left}:
    Convergence of weights in different layers on 3 datasets. The thicker the lines, the higher the layer.
    yaxis shows the $L_2$ norm of difference between the weight $w(t)$ at step $t$ with the final weight at step $t=100$.
    As expected, the fisrt layer  converges more rapidly in all three datasets. The second layer in MNIST is slightly slower than the dense layer, but it also has more pramaters. (
    For clarity, only the first 65 epochs are plotted. The architecture were all 1conv, maxpool, 1conv, maxpool, 1dense. The maxpooling are is over $2\times2$, and conv over $3\times3$ and had 32 convolutional filters).
    }
    \label{fig:converge}
\end{figure*}

Fig. \ref{fig:converge} shows empirical evidence that lower layers descend faster compared to higher layers.

\section{Lessons from Linear Regression\label{sec:reg}}

To anticipate our results about Stochastic Gradient Descent (SGD) in neural networks, let us illustrate what we would find if we were to solve linear regression using SGD.

Suppose we have a set of input vectors $X=(x_1,...,x_N)$, with $d$-dimensional $x_i$, and  $c$-dimensional labels $Y= (y_1,...,y_N)$.
We wish to find the linear transformation $A$ which minimizes the mean squared error (MSE)
\begin{equation}
    g[A] = {\mathrm{Tr}\left[HH^T\right]\over N} = {1\over N}\sum_{i=1}^N |A^Tx_i -y_i|^2 ,\quad H \equiv AX-Y
    \label{eq:greg1}
\end{equation}
The solution is
\(A = Y\tilde{X}^{-1}\)
where $X\tilde{X}^{-1} = I_d$.
Of course, in nonlinear settings, such neural networks, this solution cannot be calculated and we need to resort to methods such as SGD.
During SGD, a new batch of data $x_n$ and $y_N$ are added at each step and the estimate for $A$ is corrected based on the gradient of $g$.
%First rewrite the cost function in terms of the covariance matrix of the data
\begin{align}
    g &= \mathrm{Tr}\left[AA^TXX^T-Y^TA^TX-X^TAY+Y^TY\right]\cr
    {\delta A \over \delta N}=-\eps {\ro g\over \ro A} &= -\eps \pa{\rho A^T - XY^T} ,\qquad \rho \equiv{1\over N} XX^T
    \label{eq:reg1}
\end{align}
Note that $A$ is a matrix and all products are matrix multiplication.
$\rho$ is $d\times d$ and is similar to a covariance matrix, only without subtracting the mean.
We will refer to $\rho$ as the ``density matrix''.
We are seeking the effect of adding the $N$th data point.
Thus $A$ is evaluated using the previous $N-1$ data points.
The optimal value for $A$ at $N$ is $A_N= Y_{N-1}\tilde{X}_{N-1}^{-1}$ where $X_{N-1},Y_{N-1}$ contain only the initial $N-1$ data points and not $x_{N},y_{N}$, unlike $X=(x_1,...x_{N}),Y=(y_1,...y_{N})$ in \eqref{eq:reg1}.
We will assume the label dimension is much smaller than the input dimension, i.e. $c\ll d$.
This means that the rank of $A$ can be at most $c$ and that a right-pseudo-inverse yields $\tilde{A}_N^{-1}A_N = I'_{c,d}$ where $I'_{c,d}$ is a $d$-dimensional projection onto a $c$-dimensional subspace.
Given labels $Y_N$, we may use the pseudo-inverse $\tilde{A}_N^{-1}A_N = I'_{c,d}$
to find a ``best guess'' $\ba{X}_N$ for what the input $X_N$ might have been
\begin{align}
    \ba{X}_N\equiv \tilde{A}^{-1}_NY_N,\qquad A_N\ba{X}_{N-1} = A_NX_{N-1} = Y_{N-1} \label{eq:xtilde1}
\end{align}
although $\tilde{A}^{-1}$ and so $\ba{X}_N$ are not unique, $A_N\ba{X}_{N-1} = A_NX_{N-1} $ holds because $A_N$ is the exact solution using the $N-1$ data points.
But because $A_N$ isn't optimazed using $x_N,y_N$, $A_Nx_N \ne A_N\ba{x}_N$.
Using \eqref{eq:xtilde1} and defining $\rho_N \equiv X_NX^T_N/N$, we have
\begin{align}
    X_{N-1}\ba{X}_{N-1}^TA_N^T &= X_{N-1}X_{N-1}^TA_N^T = (N-1)\rho_{N-1} A_N^T\cr
    X_NX_N^T &= X_{N-1}X_{N-1}^T +x_Nx_N^T\cr
    \rho_N &= {N-1\over N}\rho_{N-1} +{1\over N}x_Nx_N^T
\end{align}
thus eliminating $\rho_{N-1}$ from \eqref{eq:reg1} we get
\begin{align}
    {\ro g\over \ro A_N} &= \pa{\rho_{N}  -{1\over N} X_{N}\ba{X}_{N}^T}A_N^T \cr
    & = {1\over N}x_{N}\pa{x_{N}-\ba{x}_{N}}^TA_N^T
    \label{eq:regN1}
\end{align}
Notice that $\ba{x}_N\equiv \tilde{A}_N^{-1}y_N$ is found using the $A_N$, estimated using $N-1$ data points.
Therefore, unlike \eqref{eq:xtilde1}, we have
\[\ba{x}_{N}A_{N} \ne x_{N}A_{N}\]
To better understand what \eqref{eq:regN1} means we will write $\ba{x}_N$ out explicitly. One explicit form for $\tilde{A}_N^{-1}$ is
\(\tilde{A}_N^{-1} = X_{N-1}\tilde{Y}^{-1}_{N-1}\).
Thus
\[\ba{x}_N = X_{N-1}\tilde{Y}^{-1}_{N-1} y_N \]
Writing the input $x_N$ and label $y_N$ as a linear combination of the $N-1$ previous data, we get
\begin{align}
    x_N &= X_{N-1}C,\quad
    y_N = Y_{N-1}\ba{C}, \quad  \ba{C}^T\ba{C} = 1, \quad \ba{C}\ba{C}^T = I'_{1,N} \cr
    \ba{x}_N & = X_{N-1}\ba{C},\quad \ba{x}_N\ba{x}_N^T = {1\over N-1} X_{N-1}X_{N-1}^T
    =\rho_{N-1}.
\end{align}
where $C$ is an $N\times 1 $ vector and thus $\ba{C}\ba{C}^T = I'_{1,N}$ is a rank 1 projection.
Since $x_i$ are iid, the transformation $C$ will uniformly sample previous $x_i$ and so $\ba{x}_N\ba{x}_N^T = {1\over N-1} X_{N-1}X_{N-1}^T $.
Writing $C= \ba{C}+\delta C$ and using $\ba{C}\ba{C}^T = I'_{1,N}$ yields
\begin{align}
    N\rho_N &= (N-1)\rho_{N-1} +x_Nx_N^T \cr
    &= N \rho_{N-1} + X_{N-1}\delta C x_N^T +x_N \delta C^T X_{N-1}  + O(|\delta C|^2) \cr
    & \approx N\rho_{N-1} + x_N\pa{x_N-\ba{x}_N}^T + \pa{x_N-\ba{x}_N}x_N^T
\end{align}
Thus we have
\begin{equation}
    {1\over N}x_{N}\pa{x_{N}-\ba{x}_{N}}^T = {1\over 2}\pa{\rho_N-\rho_{N-1}} = {1\over 2}{\delta \rho_N\over \delta N}\label{eq:drhodN1}
\end{equation}
Finally, we have found that SGD  \eqref{eq:regN1} is in fact relating changes in weights $A$ to changes in data covariance $\rho$ \eqref{eq:drhodN1}
\begin{equation}
    {\delta A\over \delta N} = {-\eps \over 2}{\delta \rho\over \delta N}A^T
\end{equation}
This equation can be solved by defining a ``right logarithm'' $\log_R A$ defined such that
\[d \log_R A = (dA) \tilde{A}^{-1T}\]
% $y_N = Y_{N-1}\ba{C}$, with $\ba{C}^T\ba{C} = 1$ yields $\ba{x}_N = X_{N-1}\ba{C}$

Thus, we find an equation similar what we found for the relaxation phase of a classification problem, except that it does not contain the projection on label classes $K^{(k)}_c$.
This is, of course, because there are no label classes in a regression problem.
But this also points that the linearization we used to solve the relaxation phase is implicitly assuming that the layer is stuck inside one local minimum and that the problem reduces to a set of convex optimization problems for the label classes, which explains why we were able to solve it.
The confinement to a single local minimum in the relaxation phase for the lowest layer makes sense because in this phase statistical fluctuations are not large enough to allow the layer to tunnel to a different minimum.

\outNim{
We wish to know by how much the solution for $A$ changes when a new data point $x_{N+1},y_{N+1}$ is added.

The right-pseudo-inverse of $\ba{X}^{-1}$ can be found using the singular value decomposition (SVD) of $X$.
\[X = USV^\dag, \quad \tilde{X}^{-1} = V\tilde{S}^{-1}U^\dag, \quad S\tilde{S}^{-1} = I_d\]
where $U,V$ are unitary and $S$ is a diagonal $d\times N$ matrix of the singular values (SV).

Let us interpret the solution $A^T = Y\tilde{X}^{-1}= YV\tilde{S}^{-1}U$.
The matrix $Y'=YV$ is still $c\times N$ dimensional.
}%%%%

\section{Simulations}

\section{Creating trained network layers}
Our results show that in many problems, the optimal weights in the first few layers can be found using supervised PCA.
What contrasts this from normal PCA is that each class has a different covariance (``density'') matrix and the principal components (PC) may not completely  overlap  for different classes.
%may possess a different set of principal components.
As described above, the supervised PCA consists of calculating the eigenvectors of the covariance matrix $\rho_c$ of input for each class $c$ and keeping the largest eigenvalue/ eigenvector pairs.
The number of eigenvectors to keep depends on how much of the variance of the input data we wish to keep.
Additionally, the $\rho_c$ converges very quickly and using only a fraction of the training data will still yield performance comparable to using all of the training data in many datasets.
We have experimented with different values for fraction of training data and amount variance of data used for choosing the eigenvectors, as summarized in table \ref{tab:dmn2} for MNIST and CIFAR10 datasets.

%As a simple test, we will assume that the PCs needed for all classes are the same and construct pre-trained network layers based on PCA.
We will refer to the construction described above as the Density Matrix Network (DMN).
Regular PCA has been used for creating trained layers in \cite{chan2015pcanet}, but that architecture, called PCANet, involves extra steps, which we will not have in DMN.
Even DMNs using regular PCA perform well in the first layer, but worsen in performance in higher levels (see below).
But DMNs using the class-based, supervised PCA perform within one percent of a convolutional layers (ConvNet) on MNIST \cite{MNIST}, and significantly superior to ConvNets on CIFAR10 \cite{CIFAR10}.
A single layer DMN also performs much better than a similar ConvNet on CIFAR100 \cite{CIFAR10}, and a two layer DMN is equally good as a 2 layer ConvNet.
The results are summarized in Fig. \ref{fig:results}.
Notice that it takes much less time and computation to calculate weights in a DMN as repeated backpropagation is not needed and a fraction of the training data may be sufficient.
% {\color {red} bebin inja raje be time harf mizanim bayad moghayese run time biarim. Ru hava nemishe :D}.

%The detail of the architecture of each layer will still depend on how we prepare the data.
%For instance, for images a convolutional layer is beneficial.
A DMN can also be constructed as a convolutional layer.
This is related to the Karhunen-loeve transform of images, where one breaks an image down into blocks and PCA is performed on the block images.
We only have to note that overlapping receptive fields will result in spurious PCs and we need to disentangle the outputs to get rid of them.
One simple way to do so is to use a maxpooling layer right after a DMN.
We return to this below.
The architectures we used for our experiments consist of one or two  convolutional layers with ReLU activation functions, each followed by a maxpooling layer and ending with a classification layer with softmax activation function.
We run the experiments once with DMNs for the convolutional layers and once with regular ConvNet (More experiments with hybrids of DMN and ConvNet, as well Batch Normalization \cite{ioffe2015batch} are shown in SM).

\subsection{Disentangling  the Weights using Pooling}
For images, convolutional layers work because they exploit the translational symmetry, or the fact that features can be anywhere in the image.
However, when doing PCA, we rely on the input data to have correlations only due to features intrinsic to the data.
Overlapping domains in convolution introduce spurious correlations, and hence spurious PCs, which become the most prominent PCs in PCA.
This dramatically reduces the performance of PCA, as actual features intrinsic to the data become much less significant than the spurious ones.
Thus, to correctly employ our result and use PCA to extract higher level features, the spurious correlations (``entanglements'') should be removed.
While finding the optimal way to do so may be elaborate, a simple solution is to use maxpooling. This will only keep one out of a few overlapping outputs and greatly reduce the entanglement.
therefore, we use maxpooling after every layer of DMN in all our simulations.

% \subsection{Spurious eigenvectors before disentanglement}
% {\color {red} inja ham age ye moghayese bashe bedune max pooling o ba max pooling khub mishe}.

\outNim{
\section{Rebuttal}
{\bf Rev1}:
\textbf{(i)}Sorry. Proof of Theorem3.1 follows Lemma3.2 \& 3.3 (below sec\ref{sec:th1})
\textbf{(ii)} below sec\ref{sec:SV}
\textbf{(iii)}{\bf(First)} Non-smoothness on sets of measure zero won't change result since SGD is a discrete process.
A ``smoothed'' ReLU (e.g. $x (1+ 100 \tanh x)$ works equally well for SGD.
%The smoothness is not essential for the proof.
{\bf(Second)}We should add: $w$ is on a {\em bounded domain}.
A smooth cost function with infimum but no minima attains its infimum when $|w|\to \infty$, not useful for computers simulations.
%With the bounded domain for w this is resolved.
{\bf(Third)}SGD is stochastic, perturbing $w$ randomly. So $w$ will escape unstable critical points (i.e. saddle point or maximum).
%It is stochastic by nature,
%It can only get stuck in local or global minima.
%(Note reply to  Rev.3, though. Some other points about A.2 need clarification).
\textbf{(iv)}It\^o Calculus: SGD is a stochastic equation and $N$ is the time step.
It\^o calculus makes $df/dN$ well-defined,
%Even though it looks like a finite difference equation,
%SGD can be treated as a stochastic differential equation
just as in Brownian motion, Wiener Process, etc.
\textbf{(v)}Figure3:
You are right.
The correct powers derived below (sec\ref{sec:error} \& Fig\ref{fig:1}). In eq.18 $\sqrt{\rho_c/N}$ changes to $\rho_c/ N$; PCA results won't change.

\textbf{Rev2:} We apologize for the structure (short on time).
The convergence proof relies on a separation of time-scales of layers.
It should be manifest even in shallow networks, as long as the activation function is unbounded and the initialization is reasonable.
%We agree that Fig 2 could have been done for deeper networks.
\textbf{Linearity}: Since the leading term %cannot have $b^{(k)}$ in it, since it
passes ReLU and it is not sensitive to small perturbations to $h^{(0)}$, we must have $ \prod w^{(m)} h^{(0)} \gg -b^{(k)}$.
%it still holds that, in
%{\color{red} In the relaxation phase, expanding $h^{(k)}$ in eq. 10 would yield $\prod_1^k w^{(m)} h^{(0)}$.}
\textbf{Density matrix}: Every dimension $a$ of image $i$ is a basis vector, like quantum mechanics. Each $x_i$ defines a state. $XX^T = \sum_i x_i x_i^T$ is indeed the density matrix of a mixed state.

\textbf{Rev. 3:}
\textbf{L068-L082} Can remove them. Feed-forward networks are still spin glasses, just with hierarchy.
Eergy landscape still has many local minima.
%The question of convergence is still a valid one.
%We want to emphasize that it is non-trivial to find good solutions in an energy landscape with so many local minima.
\textbf{Subsec2.3:} May be a misunderstanding; eq.2 is to blame, we will remove it.
Eq.2 is not solving SGD, neither is the loss from a single sample.
%Eq.2 shows just one of the terms in 5 and 6. %, and related to one of the indices of w and the index of b.
Eq.5 and 6 are over the whole loss.
In eq.4, 5 and 6 sum over samples $i$ is implied (will make explicit).
\textbf{Sec3:}
Sorry, the proof follows directly from Lemma 3.2, 3.3 (see below \ref{sec:th1}).
\textbf{Lemma3.2:}
We stated the reasoning incorrectly.
See sec\ref{sec:SV0}\textbf{SV} below.
\textbf{L1031-1036:}
True, it’s a bit more involved that (see below sec\ref{sec:SV}\textbf{Fast Drift \& SV:}).
\textbf{Lemma3.3:}
\textbf{(1)} see above {\bf Rev1:(iii)(First)}
\textbf{(2)} Note that your example is about the initial condition, which is arbitrary.
%It is a good example to show our argument.
The argument of Lemma 3.3 is that $h^{(k)}$ is not dominated by a single term $\prod_{m=1}^k w^{(m)} h^{(0)}$.
Starting from $h^{(k)} = \prod^k w^{(m)} h^{(0)}$ and using eq.5 and 6, after one step of SGD, %the bias becomes
%$b^{(k)} = - c {dg\over db^{(k)}} = -c {dg\over dh^{(n)}} A^{(k+1)}$.
%So, after one step
we have $h^{(k)} = \max\{0, \prod_{m=1}^k w^m h^0 - c(|h^{(k-1)}|^2+1) dg/dh^{(n)} \prod_{m=k}^n w^m\} $ ($c=$learning rate)
%which has two $w$ dependent terms, with different number of $w$’s.
Nothing guarantees that this will pass ReLU. Its $w$ dependence is unclear and it's not %, in particular it will not be
of order $\prod^k w h^0$.
\textbf{(3)} Correct. We should state: by ``A is of order B'' we mean both contain the same number of factors of $w$ (important because of Lemma 3.2).

\section{ Theorem 3.1 \label{sec:th1}}
{\bf Proof:}eq.5 has two unbounded, $k$-dependent terms: $h^{(k-1)}$ and $A^{(k+1)}$.
Lemma3.2 states $||A^{(k)}||$ is larger for smaller $k$;
Lemma3.3 that $||h^{(k-1)}|| \ll ||A^{(k+1)}h^{(0)}||$, asserting $|dg/dw^{(k)}|$ is larger for smaller $k$, QED.

{\bf Standard Error of Mean of $\rho$ \label{sec:error}}
We want fluctuations of $\rho$ with addition of data (increasing $N$).
$\rho = {1\over N}\sum_i^N x_ix_i^T$, where $x_i$ are vectors of observations of the random variable $X$ with $d$ dimensions.
For $d=1$, $\rho = \sum_i^N y_i$ with $y_i\equiv x_i^2/N$ being observations of the random variable $Y$.
Thus $ \mathrm{Var}[\rho] = N \mathrm{Var}[Y]$.
Using the Bienaym\'e formula we have $ \mathrm{Var}[Y] = \mathrm{Var}[X^2]/N^2$.  If $X$ is Gaussian with mean 0, $ \mathrm{Var}[X] = \rho $, and $\mathrm{Var}[X^2] = E[X^4] - E[X^2]^2 = 2 \mathrm{Var}[X]^2 = 2 \rho^2$.
Thus
\(\mathrm{Var}[\rho] %= 2 {\sigma_x^4\over N}
= 2 {\rho^2 \over N}\).
In other words, $\delta\rho/\delta N$ is Gaussian with mean zero and
%{%\small
%\begin{align}
$\mathrm{Var}\left[{\delta \rho\over \delta N}\right] = %{1\over N^2}
\mathrm{Var}\left[{x_N^2\over N}- {\rho\over N^2-N}\right] = {2\rho^2\over N^2} + O({1\over N^{4}}).
$ %\end{align}
%}
Thus, $\delta \rho/\delta N$ has STD $\sigma = \sqrt{2}\rho/N $.
Generalizing to $d>1$, the same equations hold for independent components of $\rho$, meaning each eigenvalue $\lambda_\mu$.
\begin{figure*}%[ht!]
    \centering
    \includegraphics[width=.5\columnwidth%,trim=0 .3cm 0 1cm
    ]{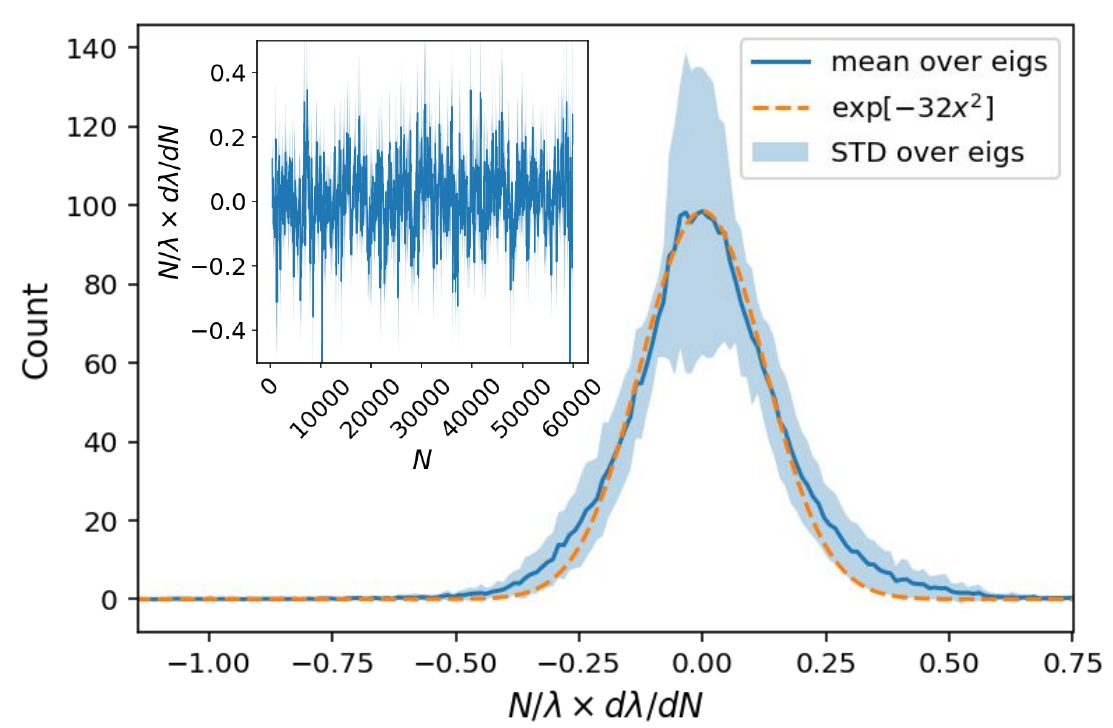}
    \caption{MNIST showing Var$[{\delta \rho \over \delta N}]\propto {\rho^2\over N^2} $ %\vspace{-15pt}
    }
    \label{fig:1}
\end{figure*}
{\bf SV: \label{sec:SV0}}
For random matrices $w$, the modulus of the largest SV is $s=\sqrt{||w||_2} \equiv  \sqrt{\mathrm{Tr}[ww^T]/d}$ with eigenvector $(1,...1)/\sqrt{d}$ ($d=$ input dim of $w$).
Writing $ww^T = s^2 I + \delta $ with $\mathrm{Tr}[\delta] =0$, we get $||\prod w^{(k)}|| = \prod {s^{(k)}}^2 $ and so the largest SV of $\prod w^{(k)}$ is $\prod s^{(k)}$.
%(mean plus fluctuations) reveals that the product  $\prod w_k = \prod s^{(k)}I + O(\delta w)$ also has an SV with modulus equal to the product of the means and uniform eigenvector.
{\bf Fast Drift \& SV: \label{sec:SV}}
The ``fast drift phase'' is where gradients ${dg\over dw}$ are larger than the ${\mathrm{STD}[w]\over \sqrt{N}}$ stochastic fluctuations.
% It does depend on the initialization.
% With the accepted Xavier initialization SV >1 will happen very early; with extremely small initialization it may not.
% Note that section 3 is about the ``Fast Drift Phase,'' which is where $||\del g||_2 > \mathrm{Var}[\delta w/\delta N] $, i.e. greater than stochastic fluctuations, which define the ``relaxation phase''.
% N is the number of samples already used and proportional to the number of steps.
At early steps, $N\sim 1$ is small, and the gradient $ - c {dg\over dw}= \Delta w > {\mathrm{STD}[w]}$. For ReLU, %a variant of
``Xavier initialization'' is used: $\sigma_{0}=\mathrm{STD}[w]=\sqrt{{2\over d}}$, %($d=$input dimension)
and largest SV of $w$ is\footnote{Since ReLU cuts the negative half, the ouput variance stays 1} 2.
%Thus we have $|\Delta w |> \sigma_{0} $.
%Like in \textbf{SV}, write $ww^T = s^2I+\delta + \Delta w \Delta w^T$.
For largest SV $s'$ of $w+\Delta w$ we have ${s'}^2 = s^2 + 2 \mathrm{Tr}[w\Delta w]/d + ||\Delta w||_2$.
$w$ has mean zero and is random, but ${dg\over dw}$, and so $\Delta w$, has a definite direction.
So the distribution of $\mathrm{Tr}[w\Delta w] $ has STD$ \sqrt{d} \sigma_0 |\Delta w|$.
Since $|\Delta w |> \sigma_{0}$, with high likelihood $ {s'}^2 > s^2 $, and we get an SV $>1$.

}
 
\section{Simple PCA layers}
Even without the supervised PCA, regular PCA can ield fairly well performing first layers.
In subsequent layrs, however, regular PCA performs poorly, mainly becuase the selected filters are not conditioned on the labels and thus may emphasize features that do not help with the classification, but rather are prominent in the data.

%As noted in the previous section, we can repeat the procedure to recursively build DMNs for higher layers, but unless we make use of $\rho_c$ to find relevant features for each class separately, the performance will get progressively worse.

To train a DMN, we simply need to find the eigenvectors of the density matrices $\rho_c$.
We will test a naive version of DMN, assuming that $\rho_c$ is is roughly independent of $c$ in the first and second layer.
This is equivalent to assuming that the low-level features contribute with similar proportions to all label classes.
This way, training a DMN is the same as doing PCA on the full $\rho = N^{-1}XX^T$.
When the the dataset is randomly sampled, $\rho$ converges very quickly (Fig. \ref{fig:rho-converge} SM).
Thus, using only a fraction of the data can give us a very good estimate for training a DMN, which can reduce the training time significantly.
Since a DMN is pretrained, the removal of one layer from backpropagation may also result in a boost in training time, especially in very deep networks.
%We test how well the average DMN (using $\rho$ instead of $\rho_c$) performs against a neural network.

\begin{figure*}
    \centering
    \includegraphics[width=.7\columnwidth, trim=2cm 0 2cm 0,clip]{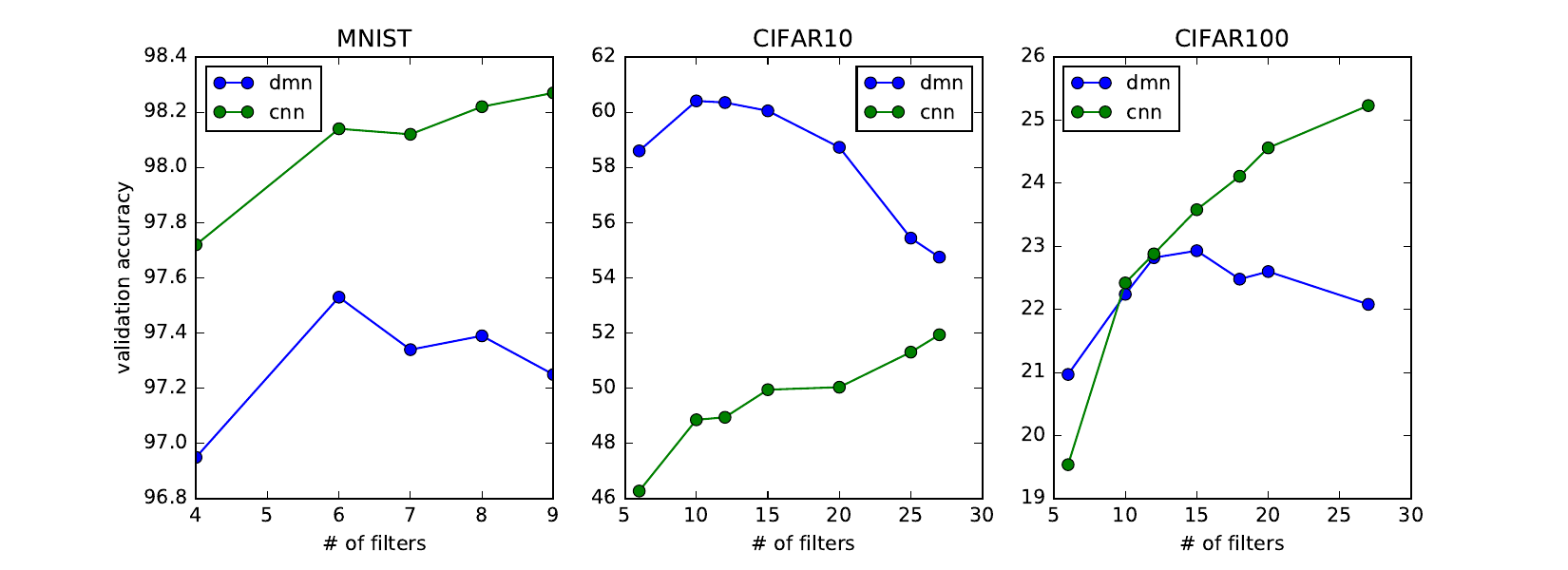}
    \caption{The comparison of test classification accuracies obtained by one dmn layer network (blue) vs one Conv layer network (green) based on the number of filters applied to MNIST, CIFAR10 and CIFAR100 datasets. The baseline models with one classification layer have the accuracies 92.9, 40.04 and 15.96 for MNIST, CIFAR10 and CIFAR100,  respectively.}
    \label{fig:1layer}
\end{figure*}

To examine the applicability of DMN layers in neural network architectures for image classification, we performed a number of tests on
%we measure the performance of networks with different setups on
three image datasets: MNIST \cite{MNIST}, CIFAR10 \cite{CIFAR10} and CIFAR100 \cite{CIFAR10}.
%We use the same training and test sets as defined in the original datasets available online \cite{MNIST}, \cite{CIFAR10} and \cite{CIFAR100}.
We simulate different scenarios and define different architectures with one or two DMNs/convolutional (Conv.) layers and compare their classification accuracies.
We use Keras \cite{chollet2015} with Tensorflow \cite{tensorflow2015-whitepaper1} back-end for all implementations.
To study the effect of DMNs and Conv. layers in the classification task, our architectures will include only one dense layer, which is the classification layer followed by a softmax activation layer.
Aside form this dense layer, the basic building block of our architectures consists of a convolutional layer, ReLU activation layer and a maxpooling layer (for disentanglement).
We use either one or two conv. layers.
Then we replace these conv. layers with DMNs and measure how the performance changes (see Fig. \ref{fig:arch} in SM for a sketch of the architectures).
%In all figures and tables demonstrating the results of the experiments,
We will denote a DMN layer with 15 filters by ``d15'', a Conv. layer with 15 filters as ``c15'', max pooling with pool size of 2 as ``m'', and dense layer with 10 nodes as ``de10''.
For DMN and Conv. layers, the kernel size has been fixed to 3.
In some experiments, we have added a batch normalization layer and we denote such layer by ``BN''.

In MNIST and CIFAR10 datasets we have 10 classes.
Therefore in the baseline model (i.e. with no conv. layers), the network consists of a single dense layer with $10$ perceptrons and softmax activation function.
For CIFAR100 model, the dense layer has $100$ perceptrons.
Next, we make the network deeper by adding one or two DMN or Conv. layers.
We test the effect of the number of filters on the performance of DMN and ConvNet.

For the first set of the experiments, which have a single DMN or Conv. layer, With one DMN, we observe that more filters does not necessarily yield better accuracy (maximum possible filters $= \textrm{(\# input channels)} \times \textrm{(kernel size)}^2$) (Fig. \ref{fig:1layer}).
However, for a network with one Conv. layer, the classification accuracy increases by the number of layers.
The key point to note here is that, DMN with maximum possible filters becomes a linear layer, whereas having less filters means we are only keeping the most prominent PCs, discarding less prominent features. for every dataset, there exists an optimal number of filters that leads to best performance of a DMN (e.g. 6 for MNIST, 12 for CIFAR10, and 15 for CIFAR100).
The results of this experiment are shown in Fig. \ref{fig:1layer}.
%As can be observed in Fig. \ref{fig:1layer}, the highest accuracy of network with one DMN layer for the MNIST dataset is obtained by 6 number of filters. In CIFAR10 and CIFAR100 datasets, 10 and 15 filters are needed to achieve the highest accuracy among other networks with one DMN layer.
Comparing the blue ``DMN'' line with green ``Conv'' line, we observe that the performance of 1 DMN layer is comparable with 1 Conv. layer.
For CIFAR10 dataset, we see that %it can be observed that for the networks with one layer of DMN/Conv
%the classification accuracy of
one DMN layer outperforms one Conv. layer.
In conclusion, we have shown that we can replace a Conv. layer with a DMN layer without degrading the classification accuracy.

In the second set of experiments we use two-layer network with different settings: 1. two DMNs, 2. one DMN and one Conv. layer and 3. two Conv. layers. Fig. \ref{fig:2layer} reports the results of two layer networks applied to three datasets.
For MNIST dataset, we can observe that adding the second DMN layer increases the classification accuracy slightly.
The performance of networks with one DMN and one Conv. layer is very close to two Conv. layer networks.
For CIFAR10 dataset,
without using BN layer, the networks with two DMN layers have lower classification accuracy compared to one DMN layer network.
However, when we add a BN layer, the classification accuracy increases slightly.
The best performance achieves by networks with one DMN and one Conv. layer and the worst performance is obtained by two Conv. layer networks.
For CIFAR100 dataset, adding the second DMN layer results in higher accuracy as can be observed from Fig. \ref{fig:2layer}.
The highest accuracy is achieved by a network composed of one DMN layer and one Conv. layer. The first layer provides a better base filters and the input data is transformed to a more meaningful feature space.
Adding a Conv. layer, back-propagation and training the network increases the classification accuracy (The classification accuracies of different architectures are summarized in tables \ref{tab:MNIST}, \ref{tab:CIFAR10}, \ref{tab:CIFAR100} in SM).
Here, we plot the effect of number of filters used in DMN layer in one layer DMN network on the classification accuracy and compare it with one Conv. layer network with the same number of filters.

\begin{figure*}
    \centering
    \includegraphics[width = .5\columnwidth]{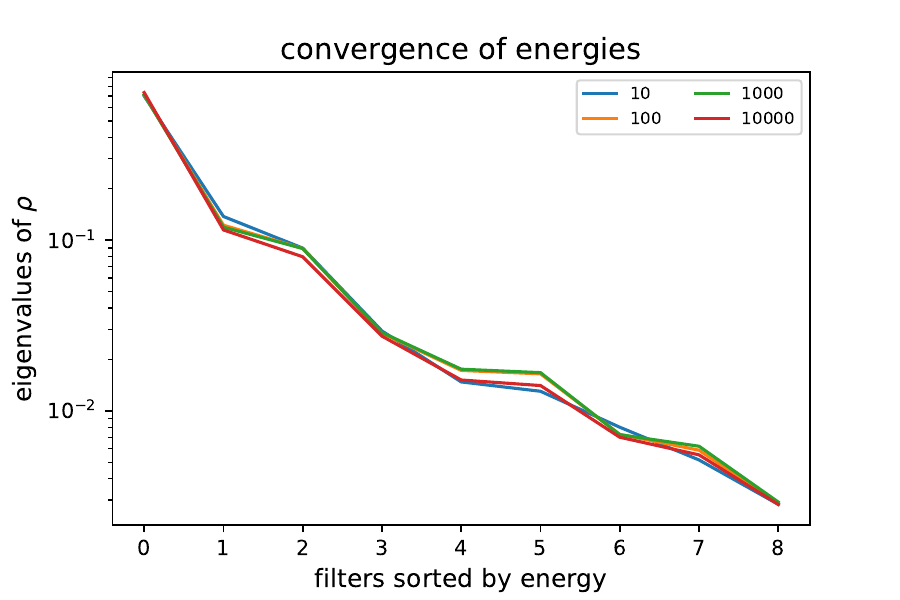}
    \caption{Convergence of eigenvalues of $\rho$, which determine the likelihood of features (lower means more likely). Using MNIST, we construct the covariance matrix $\rho$ for different number of images ($(10, 100, 1000, 10000)$ images). the images are broken down into $3\times3$ squares and flattened. The 9 eigenvalues of $\rho$ are sorted by value.
    As we see, the eigenvalues converge very quickly and a small subset of the dataset is enough for extracting these features.}
    \label{fig:rho-converge}
\end{figure*}

\section{A type of information bottleneck}
Aside from the supervised PCA,
%The key observation is that, if we are agnostic about the labels $Y$ and assume that they are diverse enough to use most features present in the data proportional to how prevalent the features are, the ideal choice for the layer weights is simply doing PCA.
%There is an important caveat here, which is the direction of our future work.
we have also shown that in cases where the classes use low-level features in a similar fashion (e.g. the MNIST dataset), one can achieve good performance using even an unsupervised PCA, agnostic to the class of the output.
In cases such as CIFAR10, however, the labels are very restrictive and each class uses low-level features in different proportions compared to the total amount features present in the data.
Moreover, the labels may pertain to rare and specific features, not necessarily the most common features in the dataset.
In these cases, simple PCA does not perform well beyond the first layer and the supervised, PCA mentioned, needs to be used to fine-tune the weights to features relevant to each class.
We have confirmed this in our simulations, shown in Figs. \ref{fig:1layer} and \ref{fig:2layer}.
We work on three datasets: MNIST, CIFAR10 and CIFAR100.
We use a naive DMN, i.e. agnostic to labels and trained using basic PCA.
In all three datasets, one layer of DMN performs at least as good as a similar convolutional layer.
We also find that, while adding
a second layer to MNIST and CIFAR100 significantly improves classification, it actually worsens the classification in CIFAR10, supporting our hypothesis that naive PCA will not help beyond first layer.

\begin{figure*}[h]
    \centering
    \includegraphics[width = .8\columnwidth]{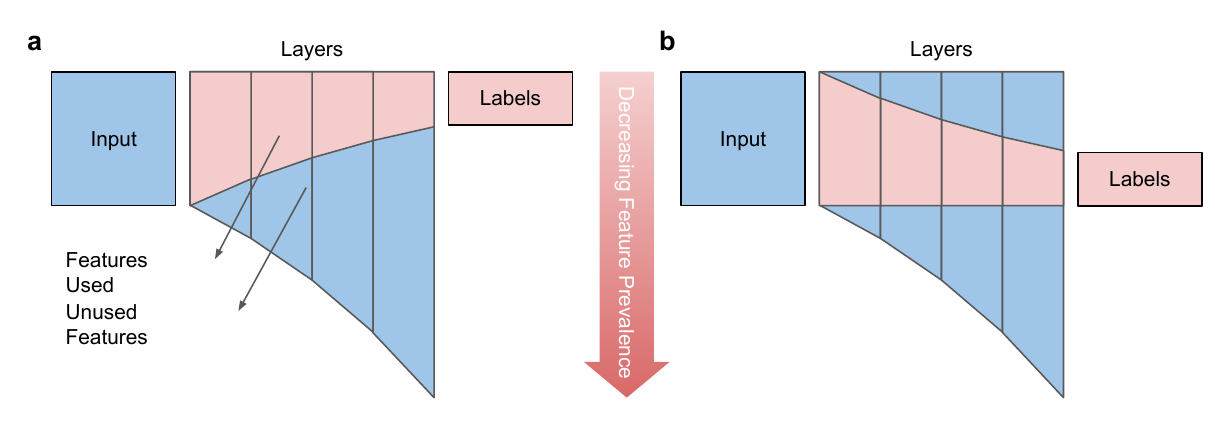}
    \caption{Growth of features, as a combination of features of the previous layer (blue columns), vs features actually useful for the desired labels (ligh red columns). The features of each column are ordered from high (top) to low in terms of their prevalence in the input data. {\bf a} An ideal case, where the input is designed to have features relevant to the desired labels as its most prevalent features (e.g. MNIST). In this case, a blind DMN with truncation of less prevalent data will lead to better classification and DMN can replace neural network layers. {\bf b} A more common and more difficult case, where the most prevalent features in the input are not useful for the classification (either not used in the desired patterns, or not helping with distinguishing between different patterns pertaining to the labels, e.g. CIFAR10 where labels are very limited, while input is very heterogeneous). In this case, Naive truncation in DMN may only work in the first layer and higher layers need a more fine-tuned approach.}
    \label{fig:bottleneck}
\end{figure*}

\section{Other simulations and test}

\begin{table}[ht]
    \centering
    {\small
    \begin{tabular}{l|c|c|c|c|c}
        {\bf MNIST}\\
        Arch. & Input & Var. & Cutoff & Val. Acc. & ConvNet Acc.\\
        \hline
d4 &0.5 &0.85 &0.9 &96.81 &97.47\\
d6 &0.3 &0.95 &0.9 &97.78 &98.12\\
d16 &0.3 &0.99 &0.9 &97.28 &98.37\\
d4, d15 &0.5 &0.85 &0.9 &97.84 &98.63\\
d6, d93 &0.3 &0.95 &0.9 &98.5 &98.86\\
\hline
    \end{tabular} \begin{tabular}{l|c|c|c|c|c}
        {\bf CIFAR10}\\
        Arch. & Input & Var. & Cutoff & Val. Acc. & ConvNet Acc.\\
        \hline
d10	&0.7 &0.99& 0.9	&49.92 & 48.61\\
d10, d11& 0.7& 0.99 &0.9& 49.94 & 46.88\\
d22 & 0.7&0.999&0.9& 56.2 & 49.94\\
d22, d41& 0.2& 0.995 &0.9& 59.64 & 51.96\\
d22, d41& 0.7& 0.995 &0.9& 59.48 & 51.96\\
d22, d87& 0.7& 0.997 &0.9& 62.15 & 52.66\\
d22, d305& 0.7& 0.999 &0.9& 63.71  & 52.13\\
\hline
    \end{tabular}
    \begin{tabular}{l|c|c|c|c|c}
        {\bf CIFAR100}\\
        Arch. & Input & Var. & Cutoff & Val. Acc. & ConvNet Acc.\\
        \hline
        d79 &1& 0.999 &0.9 &30.93&26.14\\
d79, d545 &1& 0.995 &0.9 &38.38&38.38\\
        \hline
    \end{tabular}
    }
    \caption{Comparison of DMN with ConvNet performance on MNIST, CIFAR10 and CIFAR100.
    Different thresholds for variance were used to choose the number of filters of the DMN.
    The ConvNets werte made with the same number of filters and the same architecture.
    The column ``Input'' shows the fraction of the training data used for training the last layer of  DMN compared to the ConvNet.
    As we see, even $0.2$ of the training data was enough to yield superior performance in the DMN versus ConvNet.
    }
    \label{tab:dmn2}
\end{table}

\begin{figure*}[t]
\begin{minipage}[b]{0.48\linewidth}
  \centering
  \centerline{\includegraphics[width=5.9cm]{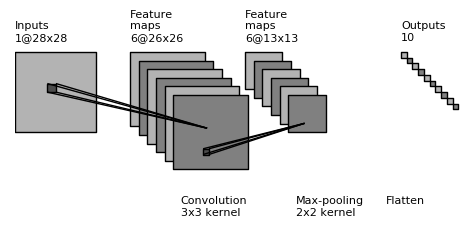}}
%  \vspace{2.0cm}
 \centerline{(a)}\medskip
\end{minipage}
%
%\hspace{0.0001cm}
%\hfill
\begin{minipage}[b]{0.48\linewidth}
  \centering
  \centerline{\includegraphics[width=8cm]{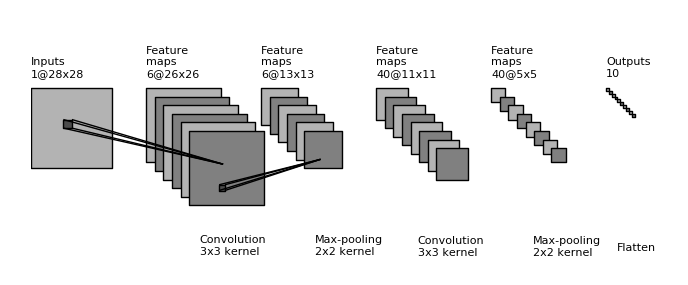}}
%  \vspace{1.5cm}
 \centerline{(b)}\medskip
\end{minipage}
\caption{
    An example of one and two layer networks used in the experiment with MNIST dataset.
    A sketch of the networks used in the experiments on MNIST, consisting of one (a) or two (b) convolutional layers with ReLU activation, each followed by a maxpooling layer and ending with a classification layer with softmax activation.
    In our experiments, we start from these architectures and then replace the convolutional layers with DMNs to compare their performances.
    The maxpooling layer is essential for the DMN as a disentangling step, making sure that the density matrix of the output does not contain spurious features arising from the overlap of the convolutional domains.
    }
\label{fig:arch}
\end{figure*}

\begin{figure*}%[ht]
    \centering
    \includegraphics[width=.8\textwidth,trim=0 .5cm 0 0 ]{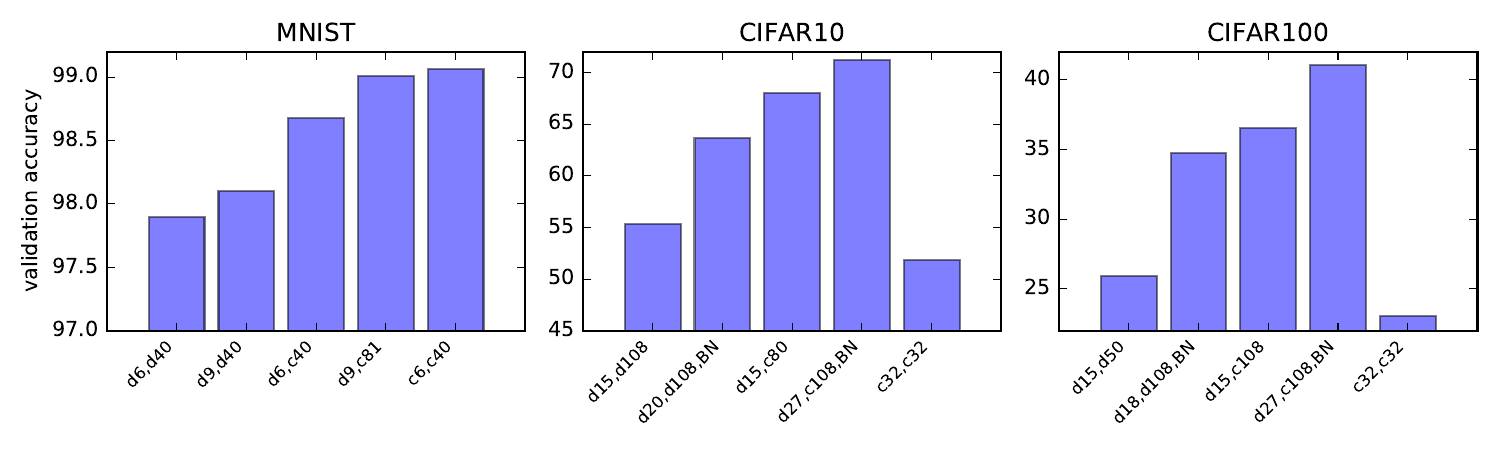}
    \caption{Test classification accuracies of different two layer networks applied to three datasets. The labels on the xaxis denote the architectures. All have 1 dense classification layer. ``d'' stands for our DMN, and ``c'' is a ConvNet. All have $3\times3$ receptive fields and after each layer is a $2\times2$ maxpool. ``d15, c80'' means 1DMN layer and one ConvNet with with 15 and 80 filters, respectively.
    \vspace{-.3cm}}
    \label{fig:2layer}
\end{figure*}

\begin{table}
    \caption{Training and test classification accuracies of different models applied to MNIST dataset. The base architecture is 1 classification layer (softmax) with 10 classes. The other layers (dmn/Conv) were added between input and the classification dense layer.}
    \label{tab:MNIST}
    \centering
    {\small
\begin{tabular}{lcc}
\hline%\toprule
 arch. &  training acc. &  validation acc. \\
\hline%\midrule
de10 &     93.51 &  92.90 \\
d6, m, de10 &   99.29 &   97.53 \\
c9, m, de10 & 99.35 &     98.27 \\
d9, m, d40, m, de10 &98.78 & 98.10 \\
d6, m, d54, m, BN, de10 &     99.90 &    98.39 \\
d9, m, BN, c81, m, de10 &   99.98 &    99.07 \\
c6, m, c40, m, de10 &  99.64 &  99.06 \\
\hline%\bottomrule
\end{tabular}
}
\end{table}

\begin{table}
    \caption{Training and test classification accuracies of different models applied to CIFAR10 dataset. The base architecture is 1 classification layer (softmax) with 10 classes. The other layers (dmn/Conv.) were added between input and the classification dense layer.}
{\small
\begin{tabular}{lcc}
\hline%\toprule
    arch. &   training acc. &  validation acc. \\
    \hline%\midrule
    de10 &     43.11 &       40.06 \\
% c48, c48, m, c96, c96, m, c192, c192, m, de512... &     0.6229 &       0.5689 \\
%  c32, c32, m, c64, c64, m, de512, de10 &    0.6866 &           0.5984 \\
    d10, m, de10 &    73.21 & 60.41 \\
    c27, m, de10 &      62.69 & 51.94 \\
    d15, m, d108, m, de10	 & 	66.11	& 55.36 \\
    d20, m, d180, m, BN, de10 &     66.02 &      63.66 \\
    d15, m, c80, m, de10 &           79.21 &      68.03 \\
    d27, m, c108, m, BN, de10 &           81.18 &           71.19 \\
    c10, m , c80, m, de10 &       58.48 &           50.10 \\
    c32, m, c32, m, de10 &       58.64 &           51.84 \\
%  d27, m, d108, m, BN, c128, m, de10 &             0.8404 &           0.6961 \\
% d20, m, c128, m, c256, m, de10 &           0.8128 &           0.7347 \\
\hline%\bottomrule
\end{tabular}
}
\label{tab:CIFAR10}
\end{table}

\begin{table}
    \caption{Training and test classification accuracies of different models applied to CIFAR100 dataset. The base architecture is 1 classification layer (softmax) with 100 classes. The other layers (dmn/Conv.) were added between input and the classification dense layer.}
{\small
\begin{tabular}{lcc}
\hline%\toprule
    arch. &  training acc. &  validation acc. \\
    \hline%\midrule
    de100 &  23.68 &     15.96 \\
 % c32, c32, m2, c64, c64, m2, de512, de10 &         0.3833 &           0.3101 \\
    d15, m, de100	&		37.48 &	22.93\\
    d18, m, BN, de100 &     51.07 &       30.30 \\
    c27, m, de100 &    41.31 & 25.23 \\
    d15, m, d50, m, de100 &     51.57 &  25.93 \\
    d18, m, d108, m, BN, de100 &      53.58 &      34.76 \\
    d15, m, c108, m, de100 &   65.06 &      36.56 \\
    d27, m, BN, c108, m, de100 &         63.57 &      41.03 \\
    c32, m, c32, m, de100 &    33.03 &     23.07 \\
    c27, m, c108, m, de100 &  36.95 &           23.97 \\
\hline%\bottomrule
\end{tabular}
}
    \label{tab:CIFAR100}
\end{table}

% \bibliography{2018/mybib}
% \bibliographystyle{plain}

\end{document}